\documentclass[lettersize,journal]{IEEEtran}
\usepackage{amsmath,amsfonts}
\usepackage{algorithmic}
\usepackage{algorithm}
\usepackage{array}
\usepackage[caption=false,font=normalsize,labelfont=sf,textfont=sf]{subfig}
\usepackage{textcomp}
\usepackage{stfloats}
\usepackage{url}
\usepackage{verbatim}
\usepackage{graphicx}
\usepackage{cite}
\usepackage{booktabs}
\usepackage{xcolor}
\usepackage{hyperref}
\usepackage{siunitx} 
\usepackage{multirow}
\usepackage{orcidlink} 
\usepackage{dsfont}

\hypersetup{colorlinks=true, 
linkcolor=blue,
citecolor=blue,
}
\newcommand{\modelname}{GVC-Seg}

\begin{document}

\title{GVC-Seg: Training-Free 3D Instance Segmentation via\\ Geometric Visual Correspondence}

\author{%
  Liang Xu\,\orcidlink{0009-0008-8302-437X}, %
  Fangjing Wang\,\orcidlink{0009-0001-0317-0617}, %
  Jinyu Yang\,\orcidlink{0009-0006-3567-6299},~\IEEEmembership{Member,~IEEE,} %
  Feng Zheng\,\orcidlink{0000-0002-1701-9141},~\IEEEmembership{Member,~IEEE}%
  \thanks{Liang Xu is with Victoria University of Wellington, New Zealand (e-mail: liang.xu@vuw.ac.nz).}%
  \thanks{Jinyu Yang is with Harbin Institute of Technology, Shenzhen, China (e-mail: jingyu.yang96@outlook.com).}%
  \thanks{Fangjing Wang and Feng Zheng are with the Department of Computer Science and Engineering, Southern University of Science and Technology, Shenzhen 518055, China (e-mail: fangjing\_wang@outlook.com; f.zheng@ieee.org).}%
  \thanks{Corresponding author: Jinyu Yang.}%
}

\markboth{Preprint submitted to IEEE}%
{Xu \MakeLowercase{\textit{et al.}}: GVC-Seg: Training-Free 3D Instance Segmentation}


\maketitle
\begin{abstract}
Accurate 3D instance segmentation in point cloud data is critical for machine vision applications. Recent advancements leverage multiple pre-trained foundation models to generate 3D proposals, followed by the application of proposal aggregation methods, which significantly enhance performance. However, they often produce sub-optimal results due to inherent variations in confidence levels across different segmentation models, resulting in a bias toward the model with higher confidence.  
This bias is inherently model-dependent and is influenced by factors such as data preprocessing techniques and training strategies.
To address this bias, we propose a novel, training-free 3D instance segmentation approach via Geometric Visual Correspondence (\modelname{}), which exploits the correspondence between 3D geometric cues and 2D visual cues to mitigate the confidence bias. 
Additionally, a 3D proposal generation module and a mask-aware CLIP feature extraction module are introduced during the instance mask generation and instance semantic reasoning, respectively. In this way, \modelname{} enhances proposal quality assessment, ensuring unbiased ensemble learning across different models.
Extensive experiments demonstrate that our method achieves state-of-the-art performance on several challenging benchmarks, while also exhibiting strong potential in open-vocabulary semantic segmentation settings.
\end{abstract}


\section{Introduction}
\label{sec:Introduction}
\IEEEPARstart{W}{ith} the advancement of depth sensing technologies, such as RGB-D cameras and LiDAR, research in 3D perception has shifted from traditional 2D image analysis to 3D point cloud processing. Among various perception tasks, 3D instance segmentation has emerged as a fundamental component of 3D scene understanding, as it enables the localization and segmentation of individual objects, which is crucial for a wide range of downstream applications~\cite{yang2020robust, liu2021scene, shan2022real, wang20233d, zhao2023lif, umam2024unsupervised, zhang2024pointgt}. However, 3D point clouds are inherently unordered and lack explicit spatial organization, making this task challenging.

The rapid progress in deep learning has spurred extensive research into the development of network architectures specifically tailored for point cloud processing~\cite{su2015multi, xu2021paconv, zhu2021cylindrical, maturana2015voxnet, qi2017pointnet, qi2017pointnetplusplus, qi2018frustum}.
A prominent approach in this area is PointNet~\cite{qi2017pointnet} and its variants~\cite{qi2017pointnetplusplus, qi2018frustum}, which utilize shared multi-layer perceptrons and max-pooling operations directly applied to point cloud data, effectively addressing the inherent challenges associated with unordered point clouds and transformation invariance. Although these models have achieved promising performance in capturing both global and local representations of point cloud targets, they still face challenges in handling issues such as stereo occlusion and significant object scale variations in real-world application scenarios.
\IEEEpubidadjcol

Initially proposed for sequential text data processing, Transformer~\cite{vaswani2017attention} has garnered considerable attention due to its order invariance and strong capability in capturing long-range dependencies. Consequently, researchers have extended its application to point cloud processing. Recent studies~\cite{zhao2021point, Schult23ICRA, ngo2023isbnet} have pioneered Transformer-based architectures for point cloud understanding, exemplified by Mask3D~\cite{Schult23ICRA} and ISBNet~\cite{ngo2023isbnet}, both of which achieve state-of-the-art results across multiple segmentation benchmarks. Despite their success, these methods strictly follow the fully supervised learning paradigm, which heavily depends on high-quality, densely annotated 3D datasets such as ScanNet~\cite{dai2017scannet, yeshwanthliu2023scannetpp, scannet200}. Since these datasets are scarce and often impractical to collect at scale, this presents a major challenge, hindering the training of scalable segmentation models and ultimately limiting the upper bound of segmentation performance in real-world applications.

To mitigate these limitations, a series of training-free paradigms~\cite{xu2023sampro3d, sam3d, ovir3d, corsetti2025functionality, nguyen2024open3dis, yin2024sai3d, openmask3d, garosi20253d, corsetti2025functionality} have been proposed. These methods aggregate proposals from pre-trained 3D or 2D foundation models, thereby enhancing segmentation performance without the need for training additional data-intensive segmentation models. Several methods~\cite{ovir3d, sam3d, xu2023sampro3d, xu2024embodieds, wang2025sgs3dhighfidelity3dinstance} leverage pre-trained 2D open-vocabulary instance segmentation networks to generate 2D proposal regions that align with textual descriptions. These regions are subsequently lifted into 3D space to derive the corresponding 3D instance masks. However, these methods face a fundamental challenge: 2D foundation models cannot fully capture the complete geometric structure of 3D objects, leading to inaccuracies in the projected 3D masks. A more recent approach~\cite{nguyen2024open3dis} employs an existing 3D instance segmentation network to generate class-agnostic instance proposals. These proposals are then merged with augmented 3D proposals, which are initially derived from an open-vocabulary 2D instance segmentation network and subsequently integrated with 3D superpoints.

\begin{figure}[t]
  \centering
  \includegraphics[width=0.98\linewidth]{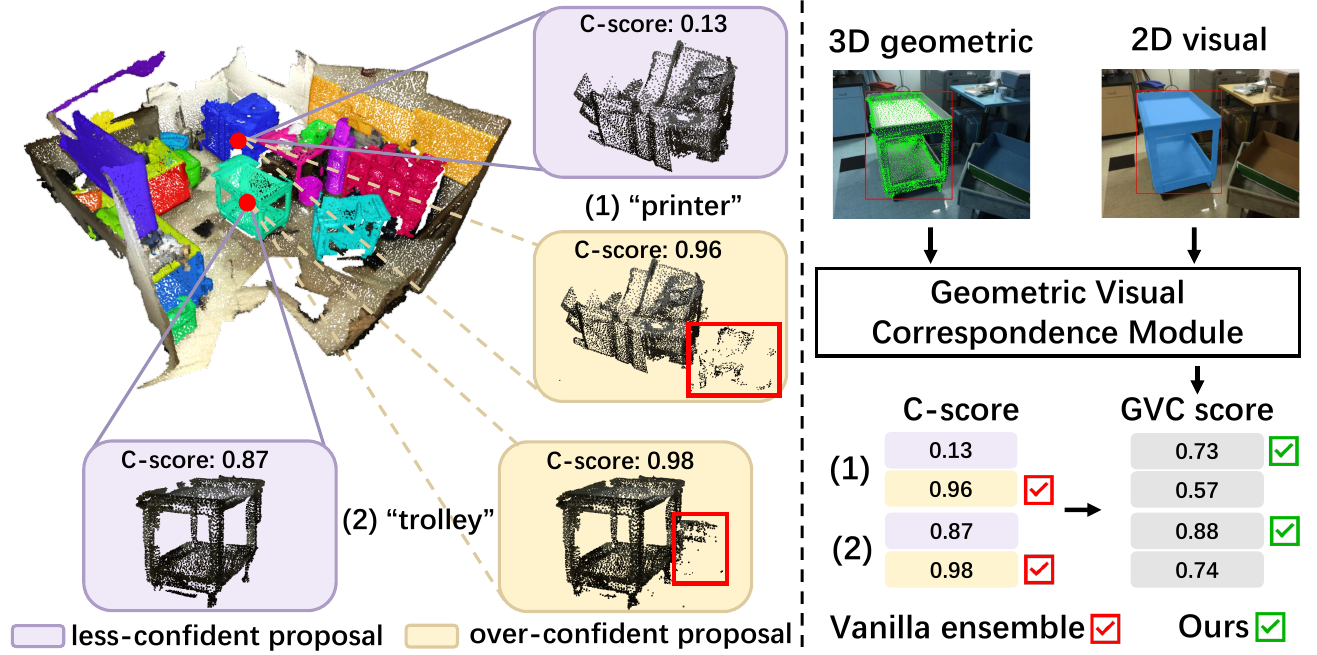}
   \caption{{Illustration of the biased confidence score (C-score) and our proposed Geometric Visual Correspondence (GVC) method. We visualize two pairs of proposals for the objects ‘printer’ and ‘trolley’. Before applying GVC, overconfident proposals are selected by conventional ensemble strategies such as Non-Maximum Suppression (NMS), leading to suboptimal segmentation performance due to their lower mask quality \textit{(defects highlighted with \textcolor{red}{red} boxes)} compared to the less-confident proposals. In contrast, GVC mitigates confidence bias and enhances segmentation results.}\label{fig:1}}
   \vspace{-15pt}
\end{figure}

However, although these training-free methods introduce more diverse 3D proposals to enhance performance, they also introduce a bias toward the more confident model when fusing proposals from different sources, as proposals generated by multiple networks exhibit varying confidence levels. {As shown in Fig. \ref{fig:1}, proposal confidences learned solely from 3D geometric structures fail to provide a reliable score for proposal quality, as the proposal with a higher mask quality is assigned a lower confidence score (less-confident proposal).}

In this work, we propose a novel 3D instance segmentation framework \modelname{}, designed to enhance segmentation performance by establishing correspondences between geometric cues from multiple 3D models and visual priors from 2D models. In particular, given a sequence of posed RGB-D images and the corresponding reconstructed 3D point cloud, \modelname{} initially employs a 3D proposal generation module to produce a diverse set of 3D object instance masks, which inherently exhibit biases due to factors such as data preprocessing methods, training strategies, and randomness during the training process. These biases can lead to a tendency for the final ensemble of proposals to favor more confident models, which does not necessarily ensure optimal segmentation results. To mitigate this bias, \modelname{} introduces a novel Geometric Visual Correspondence (GVC) confidence score for each proposal, which facilitates an unbiased NMS before extracting the mask-aware CLIP feature~\cite{radford2021learning} for semantic segmentation. Furthermore, the architecture of \modelname{} is designed to be highly extensible, allowing for the seamless integration of multiple 3D mask proposal networks within the 3D proposal generation module. This plug-and-play flexibility facilitates the incorporation of new models, thereby enhancing the robustness and quality of the generated proposals.


To assess \modelname{} on 3D instance segmentation tasks, we conduct extensive experiments on ScanNet200~\cite{scannet200}, ScanNet++\cite{yeshwanthliu2023scannetpp}, and Replica\cite{straub2019replica} datasets.
\modelname{} achieves state-of-the-art results in both closed-set and open-vocabulary settings, significantly surpassing previous methods in open-vocabulary scenarios.
Furthermore, our experiments also demonstrate the impact of scene data augmentation, proposal fusion strategies, and the choice of the number of geometric views on the segmentation performance.

Our contributions can be summarized as follows:
\begin{itemize}

\item We introduce \modelname{}, a novel training-free 3D instance segmentation framework, which focusing on reducing confidence bias while ensembling segmentation masks from different methods without additional model training.

\item We propose the Geometric Visual Correspondence (GVC) module, which integrates 3D geometric and 2D visual cues to assign an unbiased confidence score to each proposal, thereby enhancing the reliability of proposal selection and enabling more robust and unbiased outcomes.

\item Extensive experiments show that our proposed \modelname{} achieves state-of-the-art performance on ScanNet200, ScanNet++, and Replica datasets under both closed-set and open-vocabulary settings.

\end{itemize}

The rest of this paper is organized as follows. Section \ref{sec:related} presents the related works. Section \ref{sec:method} introduces the proposed GVC (Geometric Visual Correspondence) approach, which covers the proposal generation module, the implementation details of GVC, and the mask-aware CLIP feature extraction module. This is followed by experimental and ablation studies in Section \ref{sec:exp} to validate our method. Finally, we summarize the work and provide an outlook in Section \ref{sec:conclusion}.

\section{RELATED WORK}
\label{sec:related}
\subsection{Closed-Vocabulary 3D Instance Segmentation}
Instance segmentation~\cite{Schult23ICRA, ngo2023isbnet, choy20194d, fan2021scf, sun2023superpoint, wu2019pointconv, wang2019dynamic, li2018pointcnn, kundu2020virtual, kolodiazhnyi2024top} on point clouds is a crucial task in 3D scene understanding that aims to predict a unique mask for each individual object instance within a given environment, unlike semantic segmentation~\cite{graham20183d, tchapmi2017segcloud, thomas2019kpconv, zhang2023growsp, vu2023scalablesoftgroup3dinstance, wu2019pointconv, scannet200, hu2020randla}, which further assigns semantic labels to each mask.

Among the various approaches developed for 3D instance segmentation, dynamic convolution-based approaches such as Mask3D~\cite{Schult23ICRA} and ISBNet~\cite{ngo2023isbnet} have demonstrated state-of-the-art performance across multiple benchmarks. Mask3D utilizes a transformer-based architecture, representing each object through instance queries. These queries are decoded in parallel into semantic categories and instance features, enabling the generation of accurate instance masks. The transformer-based design allows Mask3D to effectively capture long-range dependencies, making it robust in complex scenes with occlusions. Furthermore, ISBNet introduces a novel enhancement by explicitly incorporating instance size as an auxiliary prior into the feature representation, enabling scale-aware differentiation. By encoding size variations within the learned features, ISBNet improves segmentation robustness, particularly for objects with significant scale differences. These advancements contribute to more accurate and adaptable 3D instance segmentation in complex real-world environments.

However, these methods can only predict categories that are predefined in the training data, which is often insufficient for real-world applications. In this paper, we leverage vision-language models to extend closed-set segmentation to novel categories.

\subsection{Open-Vocabulary 3D Instance Segmentation}
With the emergence of vision-language models~\cite{radford2021learning, pmlr-v139-jia21b, li2022languagedriven, caron2021emerging, li2019visualbert, lu2019vilbert, rao2022denseclip}, a wide range of computer vision tasks~\cite{xu2022simple, jeong2023winclip, wang2023detecting, minderer2022simple} has been endowed with open-vocabulary and zero-shot capabilities. These models generalize to unseen categories and adapt to new tasks without task-specific training, achieving promising performance.

A series of works~\cite{openmask3d, openscene, nguyen2024open3dis, huang2024openins3d} further explored the application of open-vocabulary settings in 3D scene understanding. OpenScene~\cite{openscene}, a pioneering and milestone work in applying open-vocabulary capabilities to 3D scene perception, establishes the alignment between each 3D point and the pixels of posed images in the scene. By distilling CLIP pixel features into task-agnostic point feature representations, it significantly enhances the performance of semantic segmentation on unseen categories. To further leverage the prior knowledge already encapsulated in 2D vision-language models, OpenMask3D~\cite{openmask3d} performs multi-view 2D feature fusion for each detected 3D instance mask. The fused 2D features inherently possess open-vocabulary perception capabilities, enabling retrieval with text queries of any class name, ultimately achieving open-vocabulary instance segmentation. Similarly, Open3DIS~\cite{nguyen2024open3dis} adopts this approach but further enhances it by additionally projecting 2D segmentation results into 3D space, achieving superior performance in identifying more detailed 3D proposals.

Despite significant achievements in open-vocabulary 3D instance segmentation in indoor scenes, it is worth noting that these methods face challenges in generating robust and high-quality instance proposals, which impact their overall segmentation performance. To this end, this paper proposes a dual-branch proposal generation strategy to obtain more diverse and accurate 3D instance masks.

\subsection{Segment Anything Model (SAM) in 3D}
With SAM~\cite{kirillov2023segment} bringing revolutionary advancements to 2D segmentation tasks~\cite{ding2022decoupling, li2022languagedriven, xu2022groupvit, bucher2019zero}, recent works~\cite{xu2023sampro3d, sam3d, yin2024sai3d, cen2023segment, corsetti2025functionality} have extended SAM to 3D scene understanding, enabling novel approaches for 3D segmentation. For example, OpenMask3D and Open3DIS integrate 2D SAM results to embed open-vocabulary perception capabilities within each proposal.

CIP-WPIS~\cite{yu20243d} exploits the pretrained knowledge of SAM and 3D geometric priors to generate accurate point-wise instance labels from bounding box annotations, enabling it to not only achieve state-of-the-art performance in bounding-box-supervised point cloud instance segmentation but also demonstrate robustness against noisy 3D bounding-box annotations.

On the other hand, SAMPro3D~\cite{xu2023sampro3d} seeks to identify 3D points by leveraging 3D prompts that are projected onto a sequence of posed 2D images. Initially, the pretrained SAM is applied to these 2D frames, after which it exploits the frame consistency of both the pixel-level prompts and the corresponding SAM-predicted masks to filter and consolidate the prompts. Ultimately, these refined prompts are used to derive the final segmentation masks.

Unlike these approaches, which rely on SAM’s extensive knowledge to generate additional proposals or refine visual features, in this work, we introduce multiple visual cues derived from SAM segmentation results for each 3D instance, aiming to debias the confidence between predictions from different models.

\begin{figure*}[t!]
    \centering
    \includegraphics[width=1\linewidth]{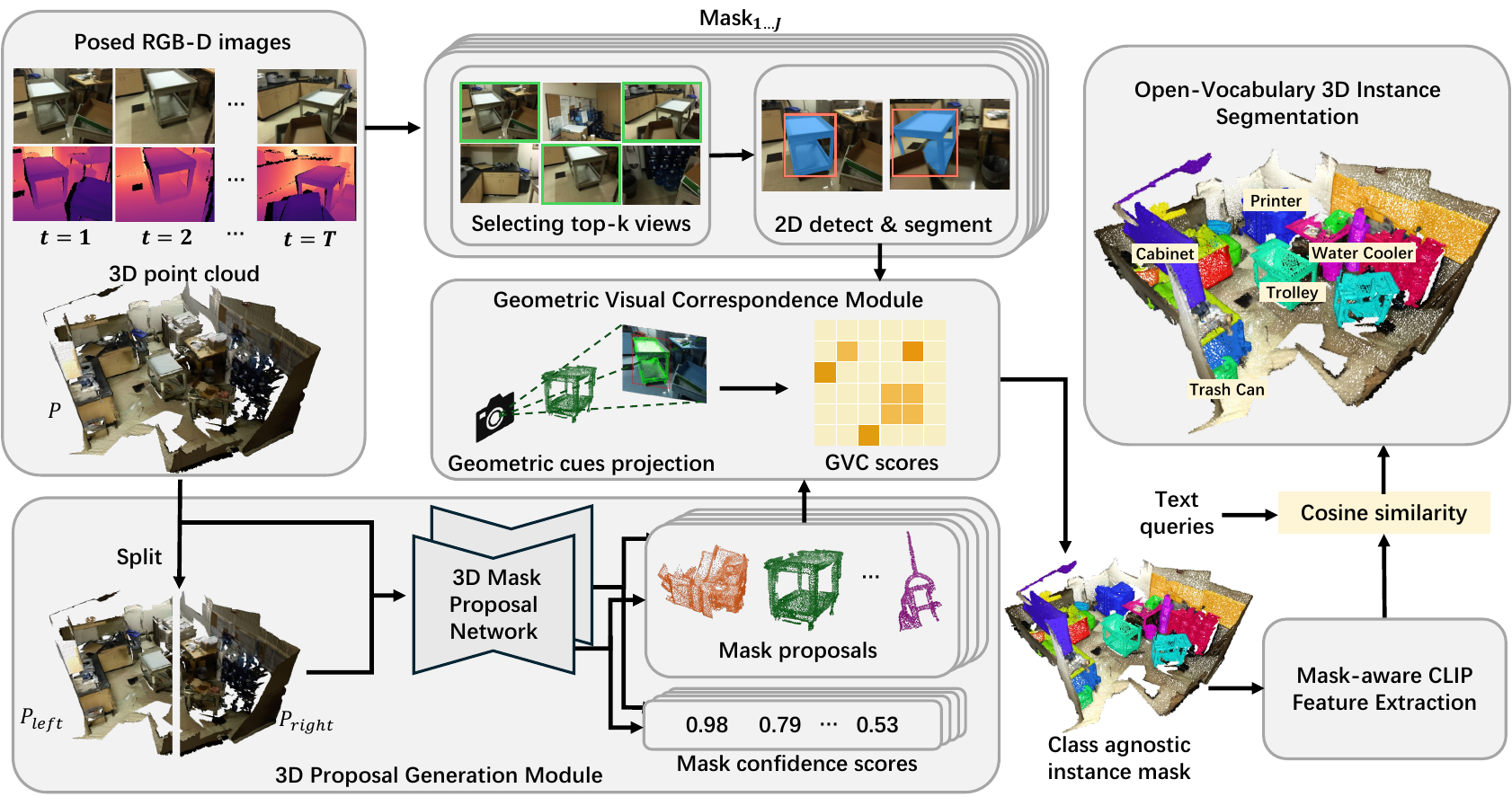}
    \caption{{Overview of our \modelname{}. The 3D Proposal Generation Module (Sec. \ref{subsec:3DPGM}) first splits the point cloud data before sending it to a dual-branch 3D proposal network to obtain abundant and biased proposal masks. These masks, along with the posed RGB-D images, are then passed to the Geometric Visual Correspondence Module (Sec. \ref{subsec:GVC Module}), where the 3D masks are projected onto 2D planes to select the most relevant views. Following this, 2D detection and segmentation are performed, and both the 3D masks and the 2D priors are used to calculate the unbiased GVC score. Finally, NMS is applied to filter redundant proposals, and per-mask CLIP features are extracted (Sec. ~\ref{subsec:featureextra}) and retrieved with category-specific text queries to enable open-vocabulary 3D instance segmentation.}}
    \label{fig:pipeline}
     \vspace{-15pt}
\end{figure*}

\section{METHODOLOGY}
\label{sec:method}

In this section, we introduce the Geometric Visual Correspondence (GVC) approach, which consists of three key modules. First, the 3D proposal generation module extracts a diverse set of biased 3D instance proposals from the input point cloud. Next, the geometric visual correspondence module establishes unbiased geometric-visual correspondences for each proposal, with the resulting unbiased scores used in the post-processing step to obtain the final unbiased predictions. Finally, the mask-aware CLIP feature extraction module generates queryable features for each instance mask, facilitating open-vocabulary 3D instance segmentation via category-specific text query retrieval.

\subsection{Overview}
Formally, our input data consists of a 3D point cloud \(\textbf{P} \in \mathbb{R}^{N \times 6}\), along with a set of posed RGB-D sequences \(\left\{I_{t}, D_{t}, C_{t}\right\}_{t=1}^{T}\), where \({I}_{t}\), \({D}_{t}\), and \({C}_{t}\) denote the RGB image, depth map, and camera parameters at each frame \(t\), respectively. We aim to predict a set of 3D object instances within the scene. Following previous methods~\cite{openmask3d, nguyen2024open3dis}, we further utilize vision-language models for instance-level semantic reasoning.

The pipeline is illustrated in Fig. \ref{fig:pipeline}. First, the proposed 3D proposal generation module processes 3D point cloud data to extract biased 3D instance masks. To improve segmentation performance on fine-grained and small-scale objects, we design a simple yet effective inference data augmentation strategy. Additionally, a dual-branch proposal network is introduced to generate more robust and diverse 3D biased proposals. Next, the geometric visual correspondence module projects each 3D segmentation result onto 2D planes and selects the most relevant views to extract 2D SAM priors, which are then utilized to compute the unbiased GVC scores. These scores incorporate auxiliary knowledge embedded in the 2D segmentation network, providing valuable priors and ensuring that the final ensembled results are both unbiased and optimal. Finally, we apply Non-Maximum Suppression (NMS) based on the GVC score, ensuring that the retained instances exhibit higher mask quality, thereby enhancing overall performance. Subsequently, we extract queryable instance representations using our mask-aware CLIP feature extraction module, enabling open-vocabulary 3D instance segmentation.


\subsection{3D Proposal Generation Module}
\label{subsec:3DPGM}
To generate class-agnostic 3D instance proposals, researchers have explored various 3D foundational models by designing new architectures and fine-tuning them with additional 3D point cloud data. In contrast to previous approaches, our proposed 3D proposal generation module aims to produce abundant biased proposals without the need for training new models. To achieve this, we first introduce a simple yet effective data augmentation strategy to enable finer-grained point cloud perception for the segmentation model. Subsequently, the augmented data is fed into pretrained 3D segmentation backbones, enhancing the detection of fine-grained instances and improving the segmentation of small-scale objects.
\subsubsection{Naive Point Cloud Data Augmentation}
Given a point cloud data \(\mathbf{P} = \left\{ \mathbf{p}_{i} \right\}_{i=1}^{N}\), where \(N\) is the number of points, and \(\mathbf{p}_{i} \in \mathbb{R}^{6}\) represents the 3D coordinates and RGB color for each point, we split the point cloud data into two smaller parts, \(\mathbf{P}_{\text{left}}\) and \(\mathbf{P}_{\text{right}}\), denoted as:
\begin{align}
    \mathbf{P}_{\text{left}} = \left\{ \mathbf{p}_{i} \right\}_{i=1}^{N}, \quad \text{where} \quad \mathbf{p}_{i,x} < \mathit{x}_{\text{mean}}, \\
    \mathbf{P}_{\text{right}} = \left\{ \mathbf{p}_{i} \right\}_{i=1}^{N}, \quad \text{where} \quad \mathbf{p}_{i,x} \geq \mathit{x}_{\text{mean}},
\end{align}
where \(x_{\text{mean}}\) denotes the mean x-axis coordinate of the points in the current scene. The point clouds \(\mathbf{P}\), \(\mathbf{P}_{\text{left}}\), and \(\mathbf{P}_{\text{right}}\) are then fed into the pre-trained 3D models for prediction. Specifically, \(\mathbf{P}_{\text{left}}\) and \(\mathbf{P}_{\text{right}}\) enable detailed instance detection and improved recognition of small-scale objects, whereas \(\mathbf{P}\) preserves a holistic understanding of the entire scene.

By employing this naive point cloud data augmentation strategy, the model is encouraged to focus more on local regions rather than the entire scene, while the input data preprocessing for the 3D segmentation networks remains unchanged. However, the splitting operation may hinder the segmentation of objects located in the central region of the scene. We leave this issue for further investigation, as the complete point cloud data \(\mathbf{P}\) is still included as input.

\subsubsection{Dual-branch Proposal Generation}
We employ two off-the-shelf 3D instance segmentation models, Mask3D and ISBNet, for proposal generation. These models process \(\mathbf{P}\), \(\mathbf{P}_{\text{left}}\), and \(\mathbf{P}_{\text{right}}\) as inputs, respectively. The segmentation outputs from these models are concatenated directly, without filtering. This strategy is deliberately adopted to enhance proposal diversity.
Assume that we obtain \(J\) instance masks \(\mathit{M} = \left\{ \mathit{m}^{3D}_{i} \right\}_{i=1}^{J}\), where each mask \(\mathit{m}^{3D}_{i}\) is a binary sequence indicating whether each 3D point belongs to the \(i\)-th instance. Alongside the masks, the 3D models also predict a confidence score for each mask, \(\mathit{S} = \left\{ \mathit{s}_{i} \right\}_{i=1}^{J}\), where \(\mathit{s}_{i} \in [0,1]\).

Notably, if an object is present in \(\mathbf{P}_{\text{left}}\), both Mask3D and ISBNet take \(\mathbf{P}_{\text{left}}\) as input and generate two distinct masks, \(\mathit{m}^{3D}_{1}\) and \(\mathit{m}^{3D}_{2}\), each associated with different confidence scores, \(\mathit{s}_{1}\) and \(\mathit{s}_{2}\), for the same object. Meanwhile, the dual-branch network processes the complete input \(\mathbf{P}\), yielding slightly different results, \(\mathit{m}^{3D}_{3}\) and \(\mathit{m}^{3D}_{4}\), along with confidence scores \(\mathit{s}_{3}\) and \(\mathit{s}_{4}\), benefiting from a broader perceptual range of the input point cloud data. Comparing \((\mathit{m}^{3D}_{1}, \mathit{s}_{1})\) with \((\mathit{m}^{3D}_{2}, \mathit{s}_{2})\), our dual-branch method introduces more robust predictions and leverages valuable priors from additional backbones. While \(\mathit{m}^{3D}_{3}\) and \(\mathit{m}^{3D}_{4}\) demonstrate superior scene-level perception capabilities, \(\mathit{m}^{3D}_{1}\) and \(\mathit{m}^{3D}_{2}\) excel in capturing fine-grained instances and small-scale objects.  

This approach significantly enriches the proposal candidate pool but also introduces biased priors that hinder ensemble performance. To mitigate this, we propose a geometric visual correspondence module to enable the unbiased selection of higher-quality candidate instances, which will be detailed in the next section.

\subsection{Geometric Visual Correspondence Module}
\label{subsec:GVC Module}
Once abundant instance proposals are generated by our proposal generation module, the next critical step is to identify and retain only those with higher mask quality. A straightforward approach is to apply Non-Maximum Suppression (NMS) to remove proposals with lower confidence scores. However, this method tends to favor proposals generated by the more confident model, without guaranteeing that higher confidence scores correspond to better mask quality. To mitigate this, we compute the correspondence between 3D geometric and 2D visual cues, providing an unbiased scoring mechanism for NMS. Specifically, we first project all 3D proposals onto 2D planes and select those views with the best visibility. We then leverage 2D foundation models on those view images to extract 2D priors, which are ultimately used to compute the unbiased correspondence score.

\subsubsection{Geometric Cues Projection}
Given the \(\mathit{i}\)-th point \(\mathbf{P}_{i} \in \mathbb{R}^{3}\) in a mask \(\mathit{m}^{3D}\), we first compute its projections onto different 2D planes, denoted as \(\mathbf{P}^{\text{image}}_{i,t} \in \mathbb{R}^{2}\), using the camera intrinsic matrix \(\mathbf{C}_{I}\) and the extrinsic matrix \(\mathbf{C}_{E_t}\). The intrinsic matrix \(\mathbf{C}_{I}\) remains constant for the same camera, whereas the extrinsic matrix \(\mathbf{C}_{E_t}\) varies with the frame index \(t\). 
\begin{align}
    h(\mathbf{p}^{\text{image}}_{i,t}) = \mathbf{C}_{I} \mathbf{C}_{E_t} h(\mathbf{p}_{i}),
\end{align}
where \(h(\cdot)\) is homogeneous coordinate transformation and \(\cdot\) indicates matrix multiplication. We then determine whether the points fall within the current camera’s field of view (FOV):
{
\begin{align}
    \mathbf{P}^{\text{image}}_{t} = \left\{ \mathbf{P}^{\text{image}}_{i,t,x,y} \right\}_{i=1}^{N}, \quad \text{if} \quad 0 < x < W \, \text{and} \, 0 < y < H,
\end{align}
}where \(W\) and \(H\) are the frame width and height, respectively.

After projecting all points in the mask \(\mathit{m}^{3D}\) onto different 2D planes using various camera extrinsic matrices \(\mathit{C}_{{E}_{t}}\), we select the top \(K\) views based on their visibility. Specifically, a projected point is considered visible if its depth coordinate (i.e., \(z_{i,t}\) in \(h(\mathbf{P}^{\text{image}}_{i,t})\)) falls within a predefined range, given by:
\begin{align}
    {\mathcal{V}}_{\text{filter}} = \mathds{1} \left( \left | z_{i,t} - d_{i,t} \right | < {\mathcal{D}}_{\text{threshold}} \right),
\end{align}
where \(\mathds{1}(\cdot)\) is an indicator function, \({\mathcal{D}}_{\text{threshold}}\) is a hyperparameter, and \(d_{i,t}\) represents the ground-truth depth value obtained from the depth map \(D\) for the \(i\)-th projected point in the \(t\)-th frame. Finally, we select the top \(K\) views that contain the highest number of visible points for each proposal.
\begin{align}
    {P}^{2D}_{\text{vis}} = \left\{ {\mathcal{V}}_{\text{filter}}(\mathbf{P}^{\text{image}}_{t}) \right\}_{t=1}^{K}.
\end{align}

\subsubsection{Geometric Visual Correspondence}
Given the top \(K\) view frames, we first extract 2D visual priors before computing the geometric-visual correspondence for each proposal. For 2D visual fundamental backbones, we leverage both detection and segmentation models. Specifically, we employ Grounding-DINO~\cite{liu2024grounding} and YOLOv9~\cite{wang2024yolov9learningwantlearn} to generate bounding boxes \(\mathit{B}^{2D} = \left\{ \mathit{b}^{2D}_{i} \right\}_{i=1}^{L}\) for the entire image, where \(L\) represents the number of detected boxes in the current frame. These boxes \(\mathit{B}^{2D}\) are then used as prompts for SAM~\cite{kirillov2023segment} to predict 2D masks \(\mathit{M}^{2D} = \left\{ \mathit{m}^{2D}_{i} \right\}_{i=1}^{L}\). The 2D detection backbone provides stronger global visual cues, while the 2D segmentation model incorporates more fine-grained pixel-level visual knowledge.
Further, we match \(\mathit{B}^{2D}, \mathit{M}^{2D}\) with projected instances \({P}^{2D}_{\text{vis}}\) by their maximum box Intersection over Union (IoU) for each view, and generate a sequence of quadruplets denoted as:
\begin{align}
    {G}_{V} = \left\{ \mathit{b}^{2D}_{i}, \mathit{b}^{2D}_{\text{vis},i}, \mathit{m}^{2D}_{i}, \mathbf{P}^{2D}_{\text{vis},i} \right\}_{i=1}^{K},
\end{align}
where \(\mathit{b}^{2D}_{\text{vis},i}\) is the minimum bounding box of the projected points \(\mathbf{P}^{2D}_{\text{vis},i}\) in the \(i\)-th view for the current proposal.

Finally, we calculate the final correspondence for a proposal with \(K\) views as:
\begin{small}
\begin{align}
    {S}_{\text{GVC}} = \frac{1}{K} \sum_{i=1}^{K} \left(
    \text{IoU}(\mathit{b}^{2D}_{i}, \mathit{b}^{2D}_{\text{vis},i})
    \times \mathcal{F}(\mathit{m}^{2D}_{i}, \mathbf{P}^{2D}_{\text{vis},i})
    \right),
\label{eq:gvc}
\end{align}
\end{small}
\begin{small}
\begin{align}
    \mathcal{F}(\mathit{m}^{2D}_{i}, \mathbf{P}^{2D}_{\text{vis},i})
    = \frac{1}{N_{i}} \sum_{x=1}^{W} \sum_{y=1}^{H}
    \mathds{1}(\mathit{m}^{2D}_{i,x,y}) \odot
    \mathds{1}(\mathbf{P}^{2D}_{\text{vis},i,x,y}),
\end{align}
\end{small} where \(N_{i}\) indicates the number of visible points in the current proposal for the \(i\)-th view, and \(\odot\) is the logical AND operation.

In this way, \(\mathit{S_{GVC}}\) incorporates auxiliary knowledge from 2D detection and segmentation networks, aligns confidence levels across proposals from different 3D segmentation backbones, and ensures an unbiased non-maximum suppression process for selecting higher-quality proposal candidates.
\begin{figure}[t]
  \centering
  \includegraphics[width=0.98\linewidth]{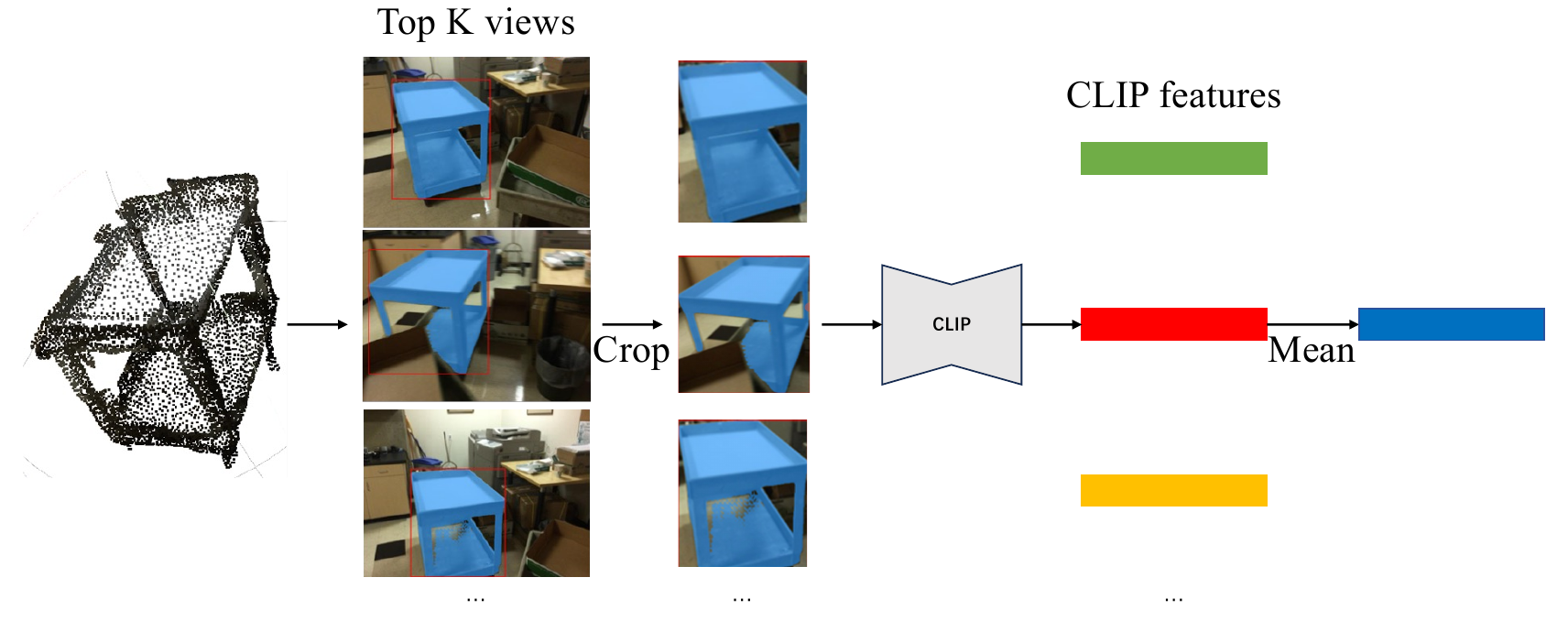}
   \caption{Illustration of the proposed Mask-aware CLIP Feature Extraction Module. The selected detection box \(\mathit{b}^{2D}_{i}\) is visualized with a red rectangle, while the mask result \(\mathit{m}^{2D}_{i}\) is displayed in blue for better clarity. Each view is cropped before being fed into the CLIP model for feature extraction. The resulting features are then averaged across all views.}\label{fig:extract}
   \vspace{-10pt}
\end{figure}

\vspace{-2pt}
\subsection{Mask-aware CLIP Feature Extraction Module}
\label{subsec:featureextra}
In the final step of our pipeline, we generate queryable feature representations for each 3D object proposal from the ensemble of proposals. This process involves utilizing the pre-computed detection results \(\mathit{b}^{2D}_{i}\) for feature extraction. Consistent with previous open-vocabulary instance segmentation methods~\cite{nguyen2024open3dis, openmask3d}, we begin by cropping the image according to each bounding box and then feed the cropped patch into the CLIP encoder. The extracted features are then aggregated by averaging across the top \(K\) views (see Fig. \ref{fig:extract}).

After extracting the CLIP feature representations for each mask, we can perform open-vocabulary semantic segmentation in a retrieval-based manner. Given a text query \( f^{\text{TEXT}} \in \mathbb{R}^{D^{\text{CLIP}}} \), we compute its similarity with the image feature \( f^{\text{IMAGE}}_{i} \in \mathbb{R}^{D^{\text{CLIP}}} \) across the top \( K \) views as follows:
{
\begin{align} 
    \mathrm{CLS} = \cos(f^{\text{TEXT}}, \frac{1}{K} \sum_{i=1}^{K} f^{\text{IMAGE}}_{i}),
\label{eq:clipquery}
\end{align}
} where \( \cos \) denotes the cosine similarity between the text and image features.

\section{EXPERIMENT}
\label{sec:exp}
To comprehensively evaluate the effectiveness and superiority of the proposed method, extensive experiments are conducted. Initially, we compare our approach with several state-of-the-art methods on multiple public benchmark datasets, including ScanNet200~\cite{scannet200}, ScanNet++~\cite{yeshwanthliu2023scannetpp}, and Replica~\cite{straub2019replica}. Following this, detailed analyses and ablation studies are provided to examine the key components and contributions of our method. Finally, we present additional qualitative results to demonstrate the performance of \modelname{} in real-world point cloud scenes, where it consistently produces higher-quality masks.

\subsection{Benchmarks and Settings}
\subsubsection{Benchmarks} We evaluate our method on three benchmarks: (i) ScanNet200\cite{scannet200}, which consists of 1,201 training and 312 validation scenes with 200 semantic categories. The evaluation is conducted on the validation set, utilizing the long-tail distribution of categories grouped into three subsets: head (66), common (68), and tail (66), based on the frequency of labeled points. This categorization makes ScanNet200 particularly suitable for assessing performance on imbalanced class distributions. Furthermore, since ScanNetV2\cite{dai2017scannet} shares identical point cloud data with ScanNet200 but only includes 20 semantic classes, we omit experiments on ScanNetV2, following the approach of Open3DIS\cite{nguyen2024open3dis}. (ii) ScanNet++\cite{yeshwanthliu2023scannetpp}, a recently introduced indoor dataset offering more detailed segmentation masks, provides a more challenging benchmark for 3D scene understanding. It comprises 280 carefully captured indoor scenes with highly accurate geometrical details and high-resolution RGB images. (iii) Replica\cite{straub2019replica}, which includes 8 evaluation scenes with 48 categories, allowing us to evaluate the generalization capability of our method. This framework addresses long-tail challenges (via ScanNet200 subsets), fine-grained understanding (via ScanNet++), and cross-dataset robustness (via Replica).
\subsubsection{Evaluation Metrics} Following previous works~\cite{yin2024sai3d,nguyen2024open3dis}, we employ Average Precision (AP) and Average Recall (AR) as our primary evaluation metrics. Specifically, we compute scores at point-level Intersection-over-Union (IoU) thresholds of 50\% and 25\% (denoted as AP@50 and AP@25, respectively), as well as the average across the IoU range of [0.5:0.95:0.05]. Additionally, we adopt two evaluation setups: class-agnostic instance segmentation and semantic instance segmentation. For ScanNet200, we further evaluate category group-specific AP, which includes three subsets: head (66 categories), common (68 categories), and tail (66 categories).

\subsubsection{Implementation Details} To generate 3D instance proposals, we employ ISBNet~\cite{ngo2023isbnet} and Mask3D~\cite{Schult23ICRA} as backbones, both of which are pretrained on the ScanNet200 dataset. For 2D image detection and segmentation, we sample every 10 frames from both RGB and depth images across all experiments. For 2D bounding box detection, we utilize the YOLOv9-E model, pretrained on the Instance-COCO dataset~\cite{lin2014microsoft}, and Grounding-DINO with the SwinT\_OGC architecture. The detected boxes are merged using Non-Maximum Suppression (NMS) with an Intersection-over-Union (IoU) threshold of 0.5. For geometric-visual correspondence, we set the number of top \(K\) views to 10, the depth threshold \({\mathcal{D}}_{threshold}\) to 0.1, and apply an NMS threshold of 0.5. For open-vocabulary (OV) semantic segmentation, we strictly adhere to the parameter settings used in OpenMask3D~\cite{openmask3d}. Notably, our method does not introduce any learnable parameters, as it is entirely training-free.

\begin{table}[t]
\caption{Class-agnostic 3D instance segmentation on ScanNet200 dataset }
    \setlength{\tabcolsep}{4pt}
    \renewcommand{\arraystretch}{1.5} 
    \centering
    \resizebox{\linewidth}{!}{
    \begin{tabular}{lcccccc}
    \toprule
    \textbf{Method} & \textbf{AP} & \textbf{AP$_{50}$} & \textbf{AP$_{25}$}  & \textbf{AR} & \textbf{AR$_{50}$} & \textbf{AR$_{25}$}  \\
    \midrule
    \text{Superpoint~\cite{sun2023superpoint}} & 5.0 & 12.7 & 38.9 & - & - & - \\
    \text{DBSCAN~\cite{ester1996density}} & 1.6 & 5.5 & 32.1 & - & - & -\\
    \text{OVIR-3D~\cite{ovir3d}} & 14.4 & 27.5 & 38.8 & - & - & -\\
    \text{Mask Clustering~\cite{yan2024maskclustering}} & 17.4 & 33.3 & 46.7 & - & - & -\\
    \text{ISBNet~\cite{ngo2023isbnet} (3D)} & 40.2 & 50.0 & 54.6 & 66.8 & 80.4 & 87.4 \\
    Open3DIS~\cite{nguyen2024open3dis} \text{ (2D+3D)} & 41.5 & 51.6 & 56.3 & \textbf{74.8} & \textbf{90.9} & \textbf{97.8} \\
    \textbf{Ours}  & \textbf{49.1} & \textbf{67.2} & \textbf{72.2} & 70.9 & 90.6 &  95.9\\
    \bottomrule
    \end{tabular}    
    }
    \label{tab:clsagnostic200}
    \par \vspace{0.5pt} 
    \hspace{-20pt}\small{The bold values mean the highest scores for the listed items.}
    \vspace{-15pt}
\end{table}

\begin{table}[!ht]
    \caption{Class-agnostic 3D instance segmentation on ScanNet++ dataset. Methods with $^{\dagger}$ are adapted. Only 3D proposal backbone methods are presented for fair comparison}
  \setlength{\tabcolsep}{6.8pt}
      \renewcommand{\arraystretch}{1.5} 
  \centering
    \begin{tabular}{lcccccc}
    \toprule
    \textbf{Method} & \multicolumn{1}{c}{\textbf{AP}} & \multicolumn{1}{c}{\textbf{AP$_{50}$}} & \multicolumn{1}{c}{\textbf{AP$_{25}$}} & \multicolumn{1}{c}{\textbf{AR}} & \multicolumn{1}{c}{\textbf{AR$_{50}$}} & \multicolumn{1}{c}{\textbf{AR$_{25}$}} \\
    \midrule
    Mask3D$^{\dagger}$\cite{Schult23ICRA} & 9.1   & 16.6  & 26.4  & 13.1  & 21.6  & 30.2 \\
    ISBNet\cite{ngo2023isbnet} & 6.2   & 10.1  & 16.2  & 10.9  & 16.9  & 25.2 \\
    \textbf{Ours (3D)} & \textbf{12.4} & \textbf{21.6} & \textbf{32.0} & \textbf{19.1} & \textbf{30.3} & \textbf{40.9} \\
    \bottomrule
    \end{tabular}%
  \label{tab:clsagnosticpp}%
    \par \vspace{5pt} 
    \hspace{-20pt}\small{The bold values mean the highest scores for the listed items.}  

    \vspace{-15pt}
\end{table}

\begin{table}[htbp]

\caption{Open-Vocabulary 3D instance segmentation (OV-3DIS) on ScanNet200 dataset}
\setlength{\tabcolsep}{5pt}
    \renewcommand{\arraystretch}{1.5} 
\centering
\begin{tabular}{lcccccc}
\toprule
\textbf{Method} & \textbf{AP} & \textbf{AP$_{50}$} & \textbf{AP$_{25}$}  & \textbf{AP}$_{\text{head}}$ & \textbf{AP}$_{\text{com}}$ & \textbf{AP}$_{\text{tail}}$ \\ 
\midrule
\multicolumn{7}{c}{Fully-supervised} \\
\midrule
ISBNet~\cite{ngo2023isbnet}  & 24.5 & 32.7 & 37.6  & 38.6 & 20.5 & 12.5  \\
Mask3D~\cite{Schult23ICRA} & \textbf{26.9} & \textbf{36.2} & \textbf{41.4} & \textbf{39.8} & \textbf{21.7} & \textbf{17.9}  \\
\midrule
\multicolumn{7}{c}{Open-vocabulary} \\
\midrule
OpenScene~\cite{openscene} & 11.7 & 15.2 & 17.8 & 13.4 & 11.6 & 9.9\\
SAM3D$^{\dagger}$~\cite{sam3d} &  6.1 & 14.2 & 21.3 & 7.0 & 6.2 & 4.6\\
OVIR-3D$^{\dagger}$~\cite{ovir3d} &  13.0 & 24.9 & 32.3 & 14.4 & 12.7 & 11.7 \\
OpenIns3D~\cite{huang2024openins3d}  & 8.8 & 10.3 & 14.4 & 16.0 & 6.5 & 4.2 \\
OpenMask3D~\cite{openmask3d} & 15.4 & 19.9 & 23.1 & 17.1 & 14.1 & 14.9 \\
Open3DIS~\cite{nguyen2024open3dis} &  23.7 & 29.4 & 32.8 & 27.8 & 21.2 & 21.8 \\
\textbf{Ours}   & \textbf{26.5} & \textbf{33.6}& \textbf{36.9} & \textbf{31.2} & \textbf{26.8} & \textbf{22.6} \\
\bottomrule
\end{tabular}
\label{tab:scannet200_OV3D}

       \vspace{-15pt}
\end{table}%

\subsection{Comparison with SOTA Methods}
\subsubsection{Fine-grained 3D Segmentation} Table \ref{tab:clsagnostic200} presents the quantitative evaluation results on the ScanNet200 dataset. We evaluate only the mask quality for the class-agnostic evaluation. Our method significantly outperforms previous approaches in terms of AP, particularly when compared to methods based on 3D backbones. Specifically, our method outperforms the baseline method, ISBNet, with improvements of +8.9 in AP and +4.1 in AR, highlighting the effectiveness of our approach in generating high-quality 3D instances. Furthermore, our method surpasses Open3DIS (2D+3D), achieving improvements of +15.6 in AP@50 and +7.6 in AP, while also achieving competitive AR scores. Through detailed analysis, we observe that the 2D branch of Open3DIS leverages SAM to generate candidate masks in the 2D image domain, resulting in a larger number of proposals and consequently higher recall. In contrast, the strength of our model lies in its ability to produce proposals with significantly higher precision.

\begin{table}[!t]
      \caption{Class-agnostic and OV-3DIS on Replica dataset}
          \renewcommand{\arraystretch}{1.5} 
  \centering
    \setlength{\tabcolsep}{5.5pt}
    \begin{tabular}{lcccccc}
    \toprule
    \multirow{2}[4]{*}{\textbf{Method}} & \multicolumn{3}{c}{\textbf{Class-agnostic}} & \multicolumn{3}{c}{\textbf{OV-3DIS}} \\
\cmidrule{2-7}          & \textbf{AP} & \textbf{AP$_{50}$} & \textbf{AP$_{25}$} & \textbf{AP} & \textbf{AP$_{50}$} & \textbf{AP$_{25}$} \\
    \midrule
    Mask3D$^{\dagger}$~\cite{Schult23ICRA} & 19.2  & 28.4  & 36.8  & 11.2  & 15.3  & 20.5 \\
    ISBNet$^{\dagger}$\cite{ngo2023isbnet} & 19.3  & 23.5  & 28.6  & 12.7  & 14.4  & 19.4 \\
    \textbf{Ours} & \textbf{24.9} & \textbf{34.9} & \textbf{43.9} & \textbf{15.2} & \textbf{18.9} & \textbf{24.8} \\
    \bottomrule
    \end{tabular}%
  \label{tab:Replica_result}%
   \vspace{-15pt}
\end{table}%

\begin{figure*}[htbp!]
    \centering
    \includegraphics[width=0.6\linewidth]{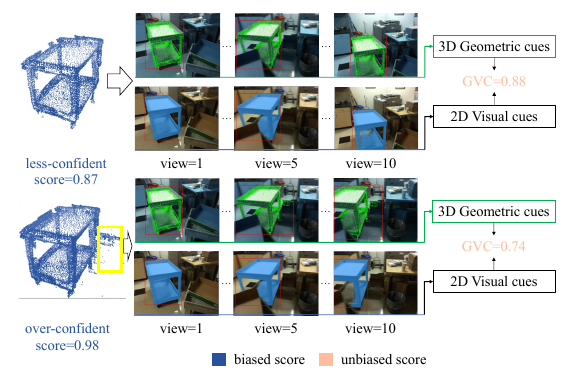}
    \caption{Visualization of a pair of predicted instance masks from different segmentation backbones. At the top, we display a less-confident proposal with a biased score of 0.87, which exhibits higher quality and is assigned a unbiased GVC score of 0.88. At the bottom, an over-confident proposal \textit{(defects marked with \textcolor{yellow}{yellow} rectangle)} with a biased score of 0.98 is adjusted to a unbiased GVC score of 0.74. For the 3D geometric cues, the projected points are visualized as green dots, and its minimum bounding rectangle is depicted as red rectangles. For the 2D visual cues, the SAM masks are displayed in blue, and the detection boxes are also represented as red rectangles.}
    \label{fig:visual_rst}
    \vspace{-15pt}
\end{figure*}
To evaluate generalizability, we conducted additional experiments on the ScanNet++ dataset. Specifically, class-agnostic 3D proposals were generated exclusively by the 3D model, intentionally excluding other 2D-to-3D lifting proposal methods to ensure a fair and unbiased comparison. As demonstrated in Table \ref{tab:clsagnosticpp}, our method outperforms previous approaches across all evaluation metrics, achieving significant improvements of +3.3 in AP and +6.0 in AR. The improvement is less pronounced compared to the results on the ScanNet200 dataset, as the backbones were not fine-tuned on ScanNet++. This supports our objective of applying existing models to new datasets in a training-free manner.

\label{subsec:ov3d}
\subsubsection{Open-Vocabulary 3D Instance Segmentation (OV-3DIS)} For open-vocabulary 3D instance segmentation evaluation, we first extract textual CLIP features for the given 200 category names in the ScanNet200 dataset and query them using per-mask features, as described in Equation \eqref{eq:clipquery}. The OV-3DIS results are presented in Table \ref{tab:scannet200_OV3D}, where our method demonstrates superior performance compared to previous state-of-the-art open-vocabulary baselines and even achieves results comparable to fully supervised methods. Specifically, our method achieves new state-of-the-art results on \(\mathit{\textbf{AP}\_{common}}\) and \(\mathit{\textbf{AP}\_{tail}}\), which represent the average precision on the common and tail subsets, respectively, highlighting its effectiveness on less common category objects.

We also conducted experiments on the Replica dataset for both class-agnostic and open-vocabulary 3D instance segmentation (OV-3DIS) tasks. As demonstrated in Table \ref{tab:Replica_result}, our method consistently outperforms the baseline models without introducing any additional training overhead. The performance improvement under the open-vocabulary setting on the Replica dataset is relatively modest, which can be attributed to the dataset's inherent simplicity, comprising only 8 scenes.

\begin{table}[!ht]
  \renewcommand{\arraystretch}{1.2} 
  \centering
  \small 
  \setlength{\tabcolsep}{5pt} 
  \caption{Ablation study on ScanNet200 for class-agnostic segmentation}
  \label{tab:ablation-semantic}
  \begin{tabular}{cccccc} 
    \toprule
    \textbf{3DPGM} & \textbf{NMS} & \textbf{GVC} & \textbf{AP} & \textbf{AP$_{50}$} & \textbf{AP$_{25}$} \\
    \midrule
    & & & 40.2 & 50.0 & 54.6 \\
    \checkmark & & & 45.0 & 56.2 & 60.8 \\
    \checkmark & \checkmark & & 45.3 & 56.8 & 61.2 \\
    \checkmark & \checkmark & \checkmark & \textbf{49.1} & \textbf{67.2} & \textbf{72.2} \\
    \bottomrule
  \end{tabular}
      \vspace{-15pt}
\end{table}

\begin{table}[!ht]
  \renewcommand{\arraystretch}{1.2} 
  \centering
  \footnotesize 
  \setlength{\tabcolsep}{3.2pt} 
  \caption{Ablation studies on ScanNet++ and Replica for class-agnostic segmentation}
  \label{tab:ablation-classagnostic}
  \begin{tabular}{ccccccccc}
    \toprule
    \multirow{2}{*}{\textbf{3DPGM}} & \multirow{2}{*}{\textbf{NMS}} & \multirow{2}{*}{\textbf{GVC}} & \multicolumn{3}{c}{\textbf{ScanNet++}} & \multicolumn{3}{c}{\textbf{Replica}} \\
    \cmidrule(lr){4-6} \cmidrule(lr){7-9}
    & & & \textbf{AP} & \textbf{AP$_{50}$} & \textbf{AP$_{25}$} & \textbf{AP} & \textbf{AP$_{50}$} & \textbf{AP$_{25}$} \\
    \midrule
    & & & 6.2 & 10.1 & 16.2 & 19.3 & 23.5 & 28.6 \\
    \checkmark & & & 8.6 & 15.5 & 25.4 & 23.5 & 28.9 & 37.1 \\
    \checkmark & \checkmark & & 8.9 & 16.2 & 26.1 & 23.9 & 29.5 & 37.8 \\
    \checkmark & \checkmark & \checkmark & \textbf{12.4} & \textbf{21.6} & \textbf{32.0} & \textbf{24.9} & \textbf{34.9} & \textbf{43.9} \\
    \bottomrule
  \end{tabular}
      \vspace{-15pt}
\end{table}

\begin{figure*}[!hbpt]
    \centering
    \includegraphics[width=1\linewidth]{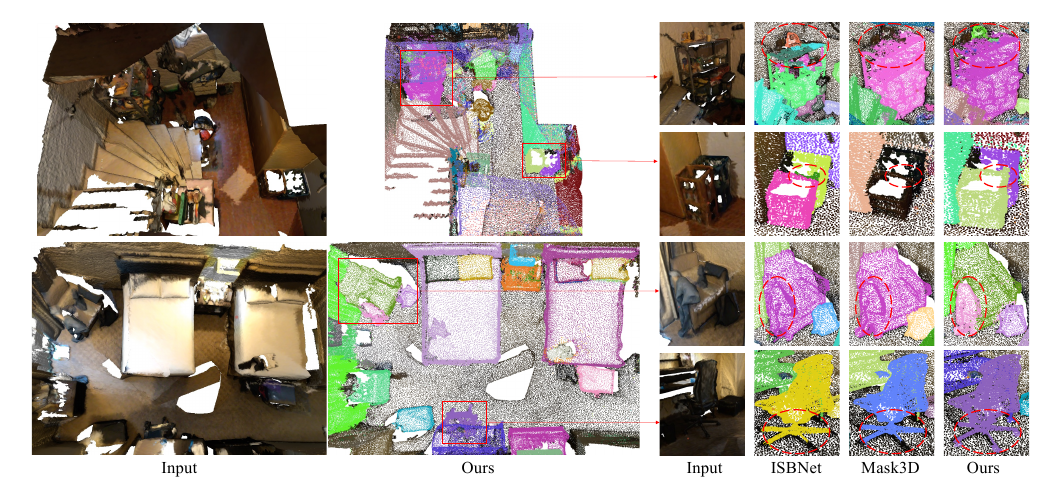}
    \caption{Qualitative results of three different methods in two different scenes. The original input data and segmentation results from ISBNet, Mask3D, and our method are shown respectively. The quality of our masks is clearly superior to the others \textit{(differences highlighted with the \textcolor{red}{red} dashed box)}. Particularly, in the third row, both ISBNet and Mask3D failed to segment the clothes on the sofa, whereas our data augmentation strategy improves the detailed segmentation.}
    \label{fig:visual_scene}
     \vspace{-15pt}
\end{figure*}

\subsection{Ablation Studies and Analysis}
\subsubsection{Effect of Each Component} Table \ref{tab:ablation-semantic} presents an ablation study validating each component on the ScanNet200 dataset. The baseline configuration, shown in the first row, exclusively uses proposals from ISBNet. We incrementally add our contributions: \text{3DPGM} refers to our proposed 3D proposal generation module, and \text{NMS} involves using Non-Maximum Suppression on the augmented proposals with the original score \(S\). In contrast, \text{GVC-NMS} utilizes our proposed geometric-visual correspondence score \({S}_{GVC}\). The results show that incorporating 3DPGM alone yields a significant +4.8 AP improvement. However, fusing additional proposals using standard NMS provides only marginal gains (+0.3 AP on ScanNet200, +0.3 AP on ScanNet++, +0.4 AP on Replica), demonstrating that naive score-based suppression is insufficient for handling the augmented proposals. Finally, integrating all components with our GVC-NMS achieves a total improvement of +8.9 AP over the baseline, with a substantial +3.8 AP gain over the NMS approach, clearly demonstrating the critical importance of our geometry-aware suppression strategy. To further evaluate the generalizability of our approach, we conducted experiments on the ScanNet++ and Replica datasets under class-agnostic settings, with results summarized in Table \ref{tab:ablation-classagnostic}. These results exhibit a similar trend of improvement as observed on the ScanNet200 dataset.

\begin{table}[!ht]
    \renewcommand{\arraystretch}{1.5} 
  \centering
      \caption{Ablation on different geometric views on ScanNet200 dataset}
  \setlength{\tabcolsep}{5.8pt}
    \begin{tabular}{ccccccc}
    \toprule
    \multicolumn{1}{c}{\textbf{Geometric views}} & \multicolumn{1}{c}{\textbf{AP}} & \multicolumn{1}{c}{\textbf{AP$_{50}$}} & \multicolumn{1}{c}{\textbf{AP$_{25}$}} & \multicolumn{1}{c}{\textbf{AR}} & \multicolumn{1}{c}{\textbf{AR$_{50}$}} & \multicolumn{1}{c}{\textbf{AR$_{25}$}} \\
    \midrule
    3     & 48.6  & 65.6  & 70.7  & 70.2  & 90.2  & 95.7 \\
    5     & 48.7  & 67.0    & 72.0  & 70.4  & 90.4  & 95.7 \\
    \textbf{10}    & \textbf{49.1} & \textbf{67.2} & \textbf{72.2}& 70.9& \textbf{90.6} & \textbf{95.9} \\
    20    & 49.0    & 65.7  & 70.6  & 71.7  & 90.3  & 95.6 \\
    100   & 48.2  & 63.9  & 70.7  & \textbf{71.8}  & 90.0    & 95.4 \\
    \bottomrule
    \end{tabular}%
  \label{tab:ablation-topk}%
  \vspace{-5pt}
\end{table}

\begin{table}[!ht]
    \renewcommand{\arraystretch}{1.2} 
    \centering
    \small 
    \setlength{\tabcolsep}{4pt} 
    \caption{Ablation on methods for correspondence calculation on ScanNet200 dataset}
    \label{tab:ablation-score}
    \begin{tabular}{lccc} 
    \toprule
    \textbf{Method} & \textbf{AP} & \textbf{AP$_{50}$} & \textbf{AP$_{25}$} \\
    \midrule
    boxes only  & 43.8 & 61.2 & 66.9 \\
    points only & 44.5 & 61.8 & 67.3 \\
    \textbf{Ours} & \textbf{49.1} & \textbf{67.2} & \textbf{72.2} \\
    \bottomrule
    \end{tabular}
    \vspace{-10pt}
\end{table}

\subsubsection{Study on the GVC Design} We analyze two key design choices in GVC: \textit{the number of geometric views} and \textit{correspondence calculation}. First, we present the performance metrics with varying numbers of geometric views on the ScanNet200 dataset, as shown in Table \ref{tab:ablation-topk}. Notably, utilizing the top 10 views achieves the best overall performance, whereas increasing the number of views to 20 or 100 results in performance degradation, likely due to the inclusion of obscured or low-quality 2D masks. Similarly, when the number of views is too small, such as with only 3 or 5 views, performance also declines due to insufficient information. Subsequently, we evaluate two methods for calculating correspondences in Table \ref{tab:ablation-score}, where we set all \(\mathcal{F}(\cdot)\) to 1 for boxes only and all \(\text{IoU}(\cdot)\) to 1 for points only in Equation \eqref{eq:gvc}. The results indicate that relying solely on bounding box IoU scores or point-level scores results in inferior performance compared to our proposed approach. This highlights the effectiveness of combining 3D geometric and 2D visual cues, where the visual cues are derived from the box-level visual cues provided by the detection model and the mask-level visual cues provided by the segmentation model.

\subsection{Qualitative Results}
Fig. \ref{fig:visual_rst} illustrates the qualitative results of our \modelname{}. We directly obtained the original confidence scores from the Mask3D and ISBNet models. Before applying GVC, the naive NMS strategy merges the two proposals on the left using the original biased confidence scores, resulting in poor segmentation performance because the proposal with a score of 0.98 has lower mask quality. In contrast, GVC-NMS mitigates this bias by selecting the higher-quality proposal with a GVC score of 0.88.

We further show the qualitative results of three different methods in two different scenes in Fig. \ref{fig:visual_scene}, where our \modelname{} consistently generates better masks. Notably, we are the only one capable of segmenting clothes on the sofa, and we also produce more accurate masks for the chair wheels.

\section{CONCLUSION}
\label{sec:conclusion}
In this paper, we introduce \modelname{}, a novel training-free 3D instance segmentation framework that leverages the correspondence between geometric cues from 3D models and visual cues from 2D models to address confidence bias in proposals, leading to significant improvements in segmentation performance. 
Extensive experiments conducted on multiple public benchmarks validate the effectiveness of \modelname{}, demonstrating its strong adaptability in the training-free setting and its capability to integrate models from diverse sources. Furthermore, we provide detailed analyses and ablation studies to investigate the key components and contributions of our approach. Qualitative results further highlight the superior performance of \modelname{} in real-world point cloud scenes. 
Looking ahead, \modelname{} has the potential to significantly enhance the performance of various downstream computer vision tasks, such as robotic navigation and manipulation, bridging the gap between performance benchmarking and real-world applications.

\bibliographystyle{IEEEtran}
\bibliography{liang}

@inproceedings{li2022languagedriven,
    title={{Language-driven Semantic Segmentation}},
    author={Boyi Li and Kilian Q Weinberger and Serge Belongie and Vladlen Koltun and Rene Ranftl},
    booktitle={Proc. ICLR},
    year={2022}
}

@inproceedings{xu2022groupvit,
  title={{GroupViT: Semantic Segmentation Emerges from Text Supervision}},
  author={Xu, Jiarui and De Mello, Shalini and Liu, Sifei and Byeon, Wonmin and Breuel, Thomas and Kautz, Jan and Wang, Xiaolong},
  booktitle={Proc. CVPR},
  pages={18134--18144},
  year={2022}
}

@INPROCEEDINGS {yan2024maskclustering,
  author = { Yan, Mi and Zhang, Jiazhao and Zhu, Yan and Wang, He },
  booktitle = {Proc. CVPR},
  title = {{ MaskClustering: View Consensus Based Mask Graph Clustering for Open-Vocabulary 3D Instance Segmentation}},
  year = {2024},
  pages = {28274-28284},
  doi = {10.1109/CVPR52733.2024.02671},
  month =Jun
}

@inproceedings{kirillov2023segment,
  title={{Segment Anything}},
  author={Kirillov, Alexander and Mintun, Eric and Ravi, Nikhila and Mao, Hanzi and Rolland, Chloe and Gustafson, Laura and Xiao, Tete and Whitehead, Spencer and Berg, Alexander C and Lo, Wan-Yen and others},
  booktitle={Proc. ICCV},
  pages={4015--4026},
  year={2023}
}

@inproceedings{ding2022decoupling,
  title={{Decoupling Zero-Shot Semantic Segmentation}},
  author={Ding, Jian and Xue, Nan and Xia, Gui-Song and Dai, Dengxin},
  booktitle={Proc. CVPR},
  pages={11583--11592},
  year={2022}
}

@inproceedings{qi2017pointnet,
  title={{PointNet: Deep Learning on Point Sets for 3D Classification and Segmentation}},
  author={Qi, Charles R and Su, Hao and Mo, Kaichun and Guibas, Leonidas J},
  booktitle={Proc. CVPR},
  pages={77--85},
  year={2017}
}

@inproceedings{qi2017pointnetplusplus,
  title     = {{PointNet++: Deep Hierarchical Feature Learning on Point Sets in a Metric Space}},
  author    = {Qi, Charles R. and Yi, Li and Su, Hao and Guibas, Leonidas J.},
  booktitle = {Proc. NeurIPS},
  volume    = {30},
  pages     = {5099--5108},
  year      = {2017}
}

@inproceedings{xu2022simple,
  title={{A Simple Baseline for Open-Vocabulary Semantic Segmentation with Pre-trained Vision-language Model}},
  author={Xu, Mengde and Zhang, Zheng and Wei, Fangyun and Lin, Yutong and Cao, Yue and Hu, Han and Bai, Xiang},
  booktitle={Proc. ECCV},
  pages={736--753},
  year={2022},
  organization={Springer}
}

@inproceedings{wang2023detecting,
  title={{Detecting Everything in the Open World: Towards Universal Object Detection}},
  author={Wang, Zhenyu and Li, Yali and Chen, Xi and Lim, Ser-Nam and Torralba, Antonio and Zhao, Hengshuang and Wang, Shengjin},
  booktitle={Proc. CVPR},
  pages={11433--11443},
  year={2023}
}

@inproceedings{yu20243d,
  title={{When 3D Bounding-Box Meets SAM: Point Cloud Instance Segmentation with Weak-and-Noisy Supervision}},
  author={Yu, Qingtao and Du, Heming and Liu, Chen and Yu, Xin},
  booktitle={Proc. WACV},
  pages={3719--3728},
  year={2024}
}

@inproceedings{minderer2022simple,
  title={{Simple Open-Vocabulary Object Detection}},
  author={Minderer, Matthias and Gritsenko, Alexey and Stone, Austin and others},
  booktitle={Proc. ECCV},
  pages={728--755},
  year={2022},
  organization={Springer}
}

@inproceedings{jeong2023winclip,
  title={{WinCLIP: Zero-/Few-Shot Anomaly Classification and Segmentation}},
  author={Jeong, Jongheon and Zou, Yang and Kim, Taewan and Zhang, Dongqing and Ravichandran, Avinash and Dabeer, Onkar},
  booktitle={Proc. CVPR},
  pages={19606--19616},
  year={2023}
}

@inproceedings{yin2024sai3d,
  title={{SAI3D: Segment Any Instance in 3D Scenes}},
  author={Yin, Yingda and Liu, Yuzheng and Xiao, Yang and Cohen-Or, Daniel and Huang, Jingwei and Chen, Baoquan},
  booktitle={Proc. CVPR},
  pages={3292--3302},
  year={2024}
}

@inproceedings{sun2023superpoint,
  title={{Superpoint Transformer for 3D Scene Instance Segmentation}},
  author={Sun, Jiahao and Qing, Chunmei and Tan, Junpeng and Xu, Xiangmin},
  booktitle={Proc. AAAI},
  volume={37},
  number={2},
  pages={2393--2401},
  year={2023}
}

@inproceedings{yeshwanthliu2023scannetpp,
  title={{{ScanNet++: A High-Fidelity Dataset of 3D Indoor Scenes}}},
  author={Yeshwanth, Chandan and Liu, Yueh-Cheng and Nie{\ss}ner, Matthias and Dai, Angela},
  booktitle = {Proc. ICCV},
  year={2023}
}

@inproceedings{xu2023sampro3d,
        title={{SAMPro3D: Locating SAM Prompts in 3D for Zero-Shot Instance Segmentation}}, 
        author={Mutian Xu and Xingyilang Yin and Lingteng Qiu and Yang Liu and Xin Tong and Xiaoguang Han},
        year={2025},
booktitle={Proc. 3DV}
  }

@inproceedings{openscene,
    title     = {OpenScene: 3D Scene Understanding with Open Vocabularies},
    author    = {Peng, Songyou and Genova, Kyle and Jiang, Chiyu "Max" and Tagliasacchi, Andrea and Pollefeys, Marc and Funkhouser, Thomas},
    booktitle = {Proc. CVPR},
    year      = {2023}
}

@inproceedings{dai2017scannet,
    title={{ScanNet: Richly-annotated 3D Reconstructions of Indoor Scenes}},
    author={Dai, Angela and Chang, Angel X. and Savva, Manolis and Halber, Maciej and Funkhouser, Thomas and Nie{\ss}ner, Matthias},
    booktitle = {Proc. CVPR},
    year = {2017}
}

@inproceedings{scannet200,
    title={{Language-Grounded Indoor 3D Semantic Segmentation in the Wild}},
    author={Rozenberszki, David and Litany, Or and Dai, Angela},
    booktitle = {Proc. ECCV},
    year={2022}
}

@inproceedings{openmask3d,
    title={{{OpenMask3D: Open-Vocabulary 3D Instance Segmentation}}},
    author={Takmaz, Ay{\c{c}}a and Fedele, Elisabetta and Sumner, Robert W. and Pollefeys, Marc and Tombari, Federico and Engelmann, Francis},
    booktitle={Proc. NeurIPS},
    year={2023}
}

@inproceedings{ovir3d,
    title={{OVIR-3D: Open-Vocabulary 3D Instance Retrieval Without Training on 3D Data}},
    author={Lu, Shiyang and Chang, Haonan and Jing, Eric Pu and Boularias, Abdeslam and Bekris, Kostas},
    booktitle={Proc. CoRL},
    year={2023}
}

@inproceedings{wu2019pointconv,
  title     = {{PointConv: Deep Convolutional Networks on 3D Point Clouds}},
  author    = {Wu, Wenxuan and Qi, Zhongang and Fuxin, Li},
  booktitle = {Proc. CVPR},
  pages     = {9621--9630},
  year      = {2019}
}

@article{vu2023scalablesoftgroup3dinstance,
  author    = {Vu, Thang and Kim, Kookhoi and Nguyen, Thanh and Luu, Tung M. and Kim, Junyeong and Yoo, Chang D.},
  journal={IEEE Trans. Pattern Anal. Mach. Intell.},
  title     = {Scalable SoftGroup for 3D Instance Segmentation on Point Clouds}, 
  year      = {2024},
  volume    = {46},
  number    = {4},
  pages     = {1981-1995},
}

@inproceedings{thomas2019kpconv,
  title={{KPConv: Flexible and Deformable Convolution for Point Clouds}},
  author={Thomas, Hugues and Qi, Charles R and Deschaud, Jean-Emmanuel and Marcotegui, Beatriz and Goulette, Fran{\c{c}}ois and Guibas, Leonidas J},
  booktitle={Proc. ICCV},
  pages={6411--6420},
  year={2019}
}

@inproceedings{ngo2023isbnet,
  title={{ISBNet: a 3D Point Cloud Instance Segmentation Network with Instance-aware Sampling and Box-aware Dynamic Convolution}},
  author={Ngo, Tuan Duc and Hua, Binh-Son and Nguyen, Khoi},
  booktitle={Proc. CVPR},
  pages={13550--13559},
  year={2023}
}

@inproceedings{Schult23ICRA,
  title     = {{Mask3D for 3D Semantic Instance Segmentation}},
  author    = {Schult, Jonas and Engelmann, Francis and Hermans, Alexander and Litany, Or and Tang, Siyu and Leibe, Bastian},
  booktitle = {Proc. ICRA},
  year      = {2023}
}

@inproceedings{caron2021emerging,
  title={{Emerging Properties in Self-Supervised Vision Transformers}},
  author={Caron, Mathilde and Touvron, Hugo and Misra, Ishan and J{\'e}gou, Herv{\'e} and Mairal, Julien and Bojanowski, Piotr and Joulin, Armand},
  booktitle={Proc. ICCV},
  pages={9650--9660},
  year={2021}
}

@inproceedings{bucher2019zero,
  title={{Zero-Shot Semantic Segmentation}},
  author={Bucher, Maxime and Vu, Tuan-Hung and Cord, Matthieu and P{\'e}rez, Patrick},
  booktitle={Proc. NeurIPS},
  volume={32},
  year={2019}
}

@inproceedings{sam3d,
  title={{SAM3D: Segment Anything in 3D Scenes}},
  author={Yang, Yunhan and Wu, Xiaoyang and He, Tong and Zhao, Hengshuang and Liu, Xihui},
  booktitle={Proc. ICCVW},
  year={2023}
}

@inproceedings{liu2024grounding,
  title={{Grounding DINO: Marrying DINO with Grounded Pre-Training for Open-Set Object Detection}},
  author={Liu, Shilong and Zeng, Zhaoyang and Ren, Tianhe and Li, Feng and Zhang, Hao and Yang, Jie and Jiang, Qing and Li, Chunyuan and Yang, Jianwei and Su, Hang and others},
  booktitle={Proc. ECCV},
  pages={38--55},
  year={2024},
  organization={Springer}
}

@inproceedings{huang2024openins3d,
      title={{OpenIns3D: Snap and Lookup for 3D Open-vocabulary Instance Segmentation}}, 
      author={Zhening Huang and Xiaoyang Wu and Xi Chen and Hengshuang Zhao and Lei Zhu and Joan Lasenby},
      booktitle={Proc. ECCV},
      year={2024}
    }

@inproceedings{radford2021learning,
  title={{Learning Transferable Visual Models From Natural Language Supervision}},
  author={Radford, Alec and Kim, Jong Wook and Hallacy, Chris and and others},
  booktitle={Proc. ICML},
  pages={8748--8763},
  year={2021},
  organization={PMLR}
}

@inproceedings{li2019visualbert,
  title={{VisualBERT: A Simple and Performant Baseline for Vision and Language}},
  author={Li, Liunian and Yatskar, Mark and Yin, Da and Hsieh, Cho-Jui and Chang, Kai-Wei},
booktitle={Proc. EMNLP},
  pages={3293--3303},
  year={2019}
}

@inproceedings{rao2022denseclip,
  title={Denseclip: Language-guided dense prediction with context-aware prompting},
  author={Rao, Yongming and Zhao, Wenliang and Chen, Guangyi and Tang, Yansong and Zhu, Zheng and Huang, Guan and Zhou, Jie and Lu, Jiwen},
  booktitle={Proc. CVPR},
  pages={18082--18091},
  year={2022}
}

@inproceedings{lu2019vilbert,
  title={{ViLBERT: Pretraining Task-Agnostic Visiolinguistic Representations for Vision-and-Language Tasks}},
  author={Lu, Jiasen and Batra, Dhruv and Parikh, Devi and Lee, Stefan},
  booktitle={Proc. NeurIPS},
  pages={13--23},
  year={2019}
}

@inproceedings{pmlr-v139-jia21b,
  title     = {Scaling Up Visual and Vision-Language Representation Learning With Noisy Text Supervision},
  author    = {Jia, Chao and Yang, Yinfei and Xia, Ye and others},
  booktitle = {Proc. ICML},
  pages     = {4904--4916},
  year      = {2021},
  volume    = {139},
  publisher = {PMLR}
}

@article{straub2019replica,
  title={{The Replica Dataset: A Digital Replica of Indoor Spaces}},
  author={Straub, Julian and Whelan, Thomas and Ma, Lingni and Chen, Yufan and Wijmans, Erik and Green, Simon and Engel, Jakob J and Mur-Artal, Raul and Ren, Carl and Verma, Shobhit and others},
  journal={arXiv preprint arXiv:1906.05797},
  year={2019}
}

@inproceedings{ester1996density,
  title={{A Density-Based Algorithm for Discovering Clusters in Large Spatial Databases with Noise}},
  author={Ester, Martin and Kriegel, Hans-Peter and Sander, J{\"o}rg and Xu, Xiaowei and others},
  booktitle={Proc. KDD},
  volume={96},
  number={34},
  pages={226--231},
  year={1996}
}

@inproceedings{graham20183d,
  title={{3D Semantic Segmentation with Submanifold Sparse Convolutional Networks}},
  author={Graham, Benjamin and Engelcke, Martin and Van Der Maaten, Laurens},
  booktitle={Proc. CVPR},
  pages={9224--9232},
  year={2018}
}

@inproceedings{lin2014microsoft,
  title={{Microsoft COCO: Common Objects in Context}},
  author={Lin, Tsung-Yi and Maire, Michael and Belongie, Serge and Hays, James and Perona, Pietro and Ramanan, Deva and Doll{\'a}r, Piotr and Zitnick, C Lawrence},
  booktitle={Proc. ECCV},
  pages={740--755},
  year={2014},
  organization={Springer}
}

@article{wang2019dynamic,
  title={{Dynamic Graph CNN for Learning on Point Clouds}},
  author={Wang, Yue and Sun, Yongbin and Liu, Ziwei and Sarma, Sanjay E and Bronstein, Michael M and Solomon, Justin M},
  journal={ACM Transactions on Graphics (TOG)},
  volume={38},
  number={5},
  pages={1--12},
  year={2019},
  publisher={Acm New York, NY, USA}
}

@inproceedings{kundu2020virtual,
  title={{Virtual Multi-view Fusion for 3D Semantic Segmentation}},
  author={Kundu, Abhijit and Yin, Xiaoqi and Fathi, Alireza and Ross, David and Brewington, Brian and Funkhouser, Thomas and Pantofaru, Caroline},
  booktitle={Proc. ECCV},
  pages={518--535},
  year={2020},
  organization={Springer}
}

@article{li2018pointcnn,
  title={{PointCNN: Convolution On X-Transformed Points}},
  author={Li, Yangyan and Bu, Rui and Sun, Mingchao and Wu, Wei and Di, Xinhan and Chen, Baoquan},
  journal={Proc. NeurIPS},
  volume={31},
  year={2018}
}

@inproceedings{hu2020randla,
  title={{RandLA-Net: Efficient Semantic Segmentation of Large-Scale Point Clouds}},
  author={Hu, Qingyong and Yang, Bo and Xie, Linhai and Rosa, Stefano and Guo, Yulan and Wang, Zhihua and Trigoni, Niki and Markham, Andrew},
  booktitle={Proc. CVPR},
  pages={11108--11117},
  year={2020}
}

@inproceedings{tchapmi2017segcloud,
  title={{SEGCloud: Semantic Segmentation of 3D Point Clouds}},
  author={Tchapmi, Lyne and Choy, Christopher and Armeni, Iro and Gwak, JunYoung and Savarese, Silvio},
booktitle={Proc. 3DV},
  pages={537--547},
  year={2017},
  organization={IEEE}
}

@inproceedings{zhang2023growsp,
  title={{GrowSP: Unsupervised Semantic Segmentation of 3D Point Clouds}},
  author={Zhang, Zihui and Yang, Bo and Wang, Bing and Li, Bo},
  booktitle={Proc. CVPR},
  pages={17619--17629},
  year={2023}
}

@inproceedings{kolodiazhnyi2024top,
  title={{Top-Down Beats Bottom-Up in 3D Instance Segmentation}},
  author={Kolodiazhnyi, Maksim and Vorontsova, Anna and Konushin, Anton and Rukhovich, Danila},
  booktitle={Proc. WACV},
  pages={3566--3574},
  year={2024}
}

@inproceedings{fan2021scf,
  title={{SCF-Net: Learning Spatial Contextual Features for Large-Scale Point Cloud Segmentation}},
  author={Fan, Siqi and Dong, Qiulei and Zhu, Fenghua and Lv, Yisheng and Ye, Peijun and Wang, Fei-Yue},
  booktitle={Proc. CVPR},
  pages={14504--14513},
  year={2021}
}

@inproceedings{choy20194d,
  title={{4D Spatio-Temporal ConvNets: Minkowski Convolutional Neural Networks}},
  author={Choy, Christopher and Gwak, JunYoung and Savarese, Silvio},
  booktitle={Proc. CVPR},
  pages={3075--3084},
  year={2019}
}

@article{cen2023segment,
  title={{Segment anything in 3D with NeRFs}},
  author={Cen, Jiazhong and Zhou, Zanwei and Fang, Jiemin and Shen, Wei and Xie, Lingxi and Jiang, Dongsheng and Zhang, Xiaopeng and Tian, Qi and others},
  journal={Proc. NeurIPS},
  volume={36},
  pages={25971--25990},
  year={2023}
}

@inproceedings{nguyen2024open3dis,
  title={{Open3DIS: Open-Vocabulary 3D Instance Segmentation with 2D Mask Guidance}},
  author={Nguyen, Phuc and Ngo, Tuan Duc and Kalogerakis, Evangelos and Gan, Chuang and Tran, Anh and Pham, Cuong and Nguyen, Khoi},
  booktitle={Proc. CVPR},
  pages={4018--4028},
  year={2024}
}

@inproceedings{wang2024yolov9learningwantlearn,
  title={Yolov9: Learning what you want to learn using programmable gradient information},
  author={Wang, Chien-Yao and Yeh, I-Hau and Mark Liao, Hong-Yuan},
  booktitle={Proc. ECCV},
  pages={1--21},
  year={2024},
  organization={Springer}
}

@article{yang2020robust,
  title={{Robust and Efficient RGB-D SLAM in Dynamic Environments}},
  author={Yang, Xin and Yuan, Zikang and Zhu, Dongfu and Chi, Cheng and Li, Kun and Liao, Chunyuan},
  journal={IEEE Transactions on Multimedia (TMM)},
  volume={23},
  pages={4208--4219},
  year={2020},
  publisher={IEEE}
}

@article{liu2021scene,
  title={{Scene Recognition Mechanism for Service Robot Adapting Various Families: A CNN-Based Approach Using Multi-Type Cameras}},
  author={Liu, Shaopeng and Tian, Guohui and Zhang, Ying and Duan, Peng},
  journal={IEEE Transactions on Multimedia (TMM)},
  volume={24},
  pages={2392--2406},
  year={2021},
  publisher={IEEE}
}

@article{shan2022real,
  title={{Real-time 3D Single Object Tracking with Transformer}},
  author={Shan, Jiayao and Zhou, Sifan and Cui, Yubo and Fang, Zheng},
  journal={IEEE Transactions on Multimedia (TMM)},
  volume={25},
  pages={2339--2353},
  year={2022},
  publisher={IEEE}
}

@article{umam2024unsupervised,
  title={{Unsupervised Point Cloud Co-part Segmentation via Co-attended Superpoint Generation and Aggregation}},
  author={Umam, Ardian and Yang, Cheng-Kun and Chuang, Jen-Hui and Lin, Yen-Yu},
  journal={IEEE Transactions on Multimedia (TMM)},
  year={2024},
  publisher={IEEE}
}

@article{wang20233d,
  title={{3D Object Segmentation using Cross-Window Point Transformer with Latent Semantic Boundary Guidance}},
  author={Wang, Qide and Liu, Daxin and Liu, Zhenyu and Xu, Jiatong and Tan, Jianrong},
  journal={IEEE Transactions on Multimedia (TMM)},
  year={2023},
  publisher={IEEE}
}

@article{zhang2024pointgt,
  title={{PointGT: A Method for Point-Cloud Classification and Segmentation Based on Local Geometric Transformation}},
  author={Zhang, Huang and Wang, Changshuo and Yu, Long and Tian, Shengwei and Ning, Xin and Rodrigues, Joel},
  journal={IEEE Transactions on Multimedia (TMM)},
  year={2024},
  publisher={IEEE}
}

@article{zhao2023lif,
  title={{LIF-Seg: LiDAR and Camera Image Fusion for 3D LiDAR Semantic Segmentation}},
  author={Zhao, Lin and Zhou, Hui and Zhu, Xinge and Song, Xiao and Li, Hongsheng and Tao, Wenbing},
  journal={IEEE Transactions on Multimedia (TMM)},
  volume={26},
  pages={1158--1168},
  year={2023},
  publisher={IEEE}
}

@article{zhu2021cylindrical,
  title={{Cylindrical and Asymmetrical 3D Convolution Networks for LiDAR-based Perception}},
  author={Zhu, Xinge and Zhou, Hui and Wang, Tai and Hong, Fangzhou and Li, Wei and Ma, Yuexin and Li, Hongsheng and Yang, Ruigang and Lin, Dahua},
  journal={IEEE Trans. Pattern Anal. Mach. Intell.},
  volume={44},
  number={10},
  pages={6807--6822},
  year={2021},
  publisher={IEEE}
}

@inproceedings{xu2021paconv,
  title={{PAConv: Position Adaptive Convolution with Dynamic Kernel Assembling on Point Clouds}},
  author={Xu, Mutian and Ding, Runyu and Zhao, Hengshuang and Qi, Xiaojuan},
  booktitle={Proc. CVPR},
  pages={3173--3182},
  year={2021}
}

@inproceedings{maturana2015voxnet,
  title={{VoxNet: A 3D Convolutional Neural Network for real-time object recognition}},
  author={Maturana, Daniel and Scherer, Sebastian},
  booktitle={Proc. IROS},
  pages={922--928},
  year={2015},
  organization={IEEE}
}

@inproceedings{su2015multi,
  title={{Multi-view Convolutional Neural Networks for 3D Shape Recognition}},
  author={Su, Hang and Maji, Subhransu and Kalogerakis, Evangelos and Learned-Miller, Erik},
  booktitle={Proc. ICCV},
  pages={945--953},
  year={2015}
}

@article{vaswani2017attention,
  title={{Attention Is All You Need}},
  author={Vaswani, Ashish and Shazeer, Noam and Parmar, Niki and Uszkoreit, Jakob and Jones, Llion and Gomez, Aidan N and Kaiser, {\L}ukasz and Polosukhin, Illia},
  journal={Proc. NeurIPS},
  volume={30},
  year={2017}
}

@inproceedings{zhao2021point,
  title={{Point Transformer}},
  author={Zhao, Hengshuang and Jiang, Li and Jia, Jiaya and Torr, Philip HS and Koltun, Vladlen},
  booktitle={Proc. ICCV},
  pages={16259--16268},
  year={2021}
}

@inproceedings{qi2018frustum,
  title={{Frustum PointNets for 3D Object Detection from RGB-D Data}},
  author={Qi, Charles R and Liu, Wei and Wu, Chenxia and Su, Hao and Guibas, Leonidas J},
  booktitle={Proc. CVPR},
  pages={918--927},
  year={2018}
}

@inproceedings{wang2025sgs3dhighfidelity3dinstance,
  title     = {{SGS-3D: High-Fidelity 3D Instance Segmentation via Reliable Semantic Mask Splitting and Growing}},
  author    = {Wang, Chaolei and Luo, Yang and Du, Jing and Chen, Siyu and Chen, Yiping and Han, Ting},
  booktitle={Proc. AAAI},
  year      = {2025}
}

@inproceedings{xu2024embodieds,
  title={{EmbodiedSAM: Online Segment Any 3D Thing in Real Time}},
  author={Xu, Xiuwei and Chen, Huangxing and Zhao, Linqing and Wang, Ziwei and Zhou, Jie and Lu, Jiwen},
  booktitle={Proc. ICLR},
  year={2025}
}

@inproceedings{corsetti2025functionality,
  title={{Functionality Understanding and Segmentation in 3D Scenes}},
  author={Corsetti, Jaime and Giuliari, Francesco and Fasoli, Alice and Boscaini, Davide and Poiesi, Fabio},
  booktitle={Proc. CVPR},
  pages={24550--24559},
  year={2025}
}

@inproceedings{garosi20253d,
  title={{3D Part Segmentation via Geometric Aggregation of 2D Visual Features}},
  author={Garosi, Marco and Tedoldi, Riccardo and Boscaini, Davide and Mancini, Massimiliano and Sebe, Nicu and Poiesi, Fabio},
  booktitle={Proc. WACV},
  pages={3257--3267},
  year={2025},
  organization={IEEE}
}
\begin{IEEEbiography}
[{\includegraphics[width=1in,height=1.25in,clip,keepaspectratio]{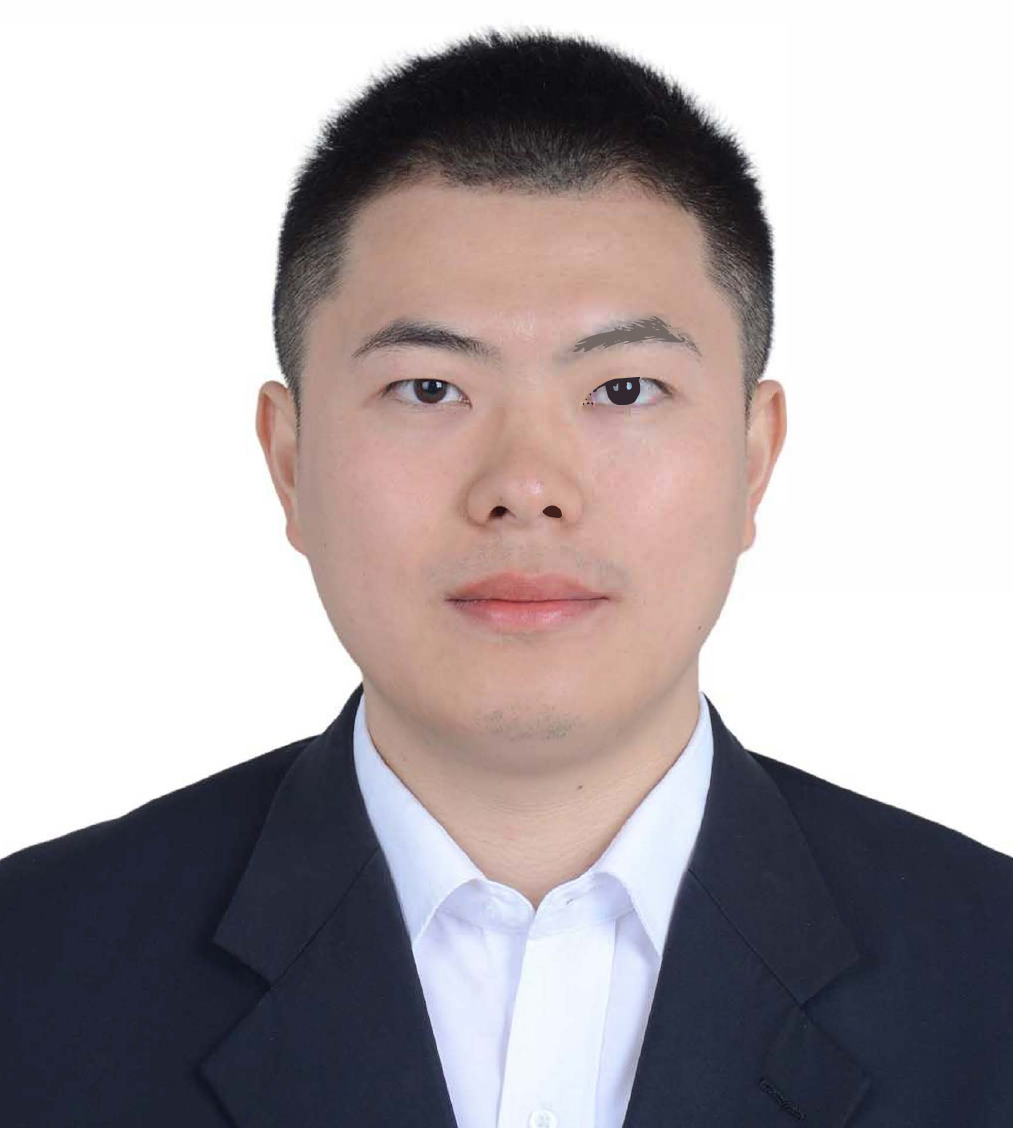}}]{Liang Xu}
received his Bachelor's degree in Biomedical Engineering from Xidian University, Xi'an, China, in 2016, and his M.S. degree in Information and Communication Engineering from Huazhong University of Science and Technology (HUST), Wuhan, China, in 2019. He is currently pursuing a Ph.D. degree in the School of Engineering and Computer Science, Victoria University of Wellington, Wellington 6140,
New Zealand. His research interests include computer vision, speech signal processing, and deep learning.
\end{IEEEbiography}
\vspace{-30pt}
\begin{IEEEbiography}[{\includegraphics[width=1in,height=1.25in,clip,keepaspectratio]{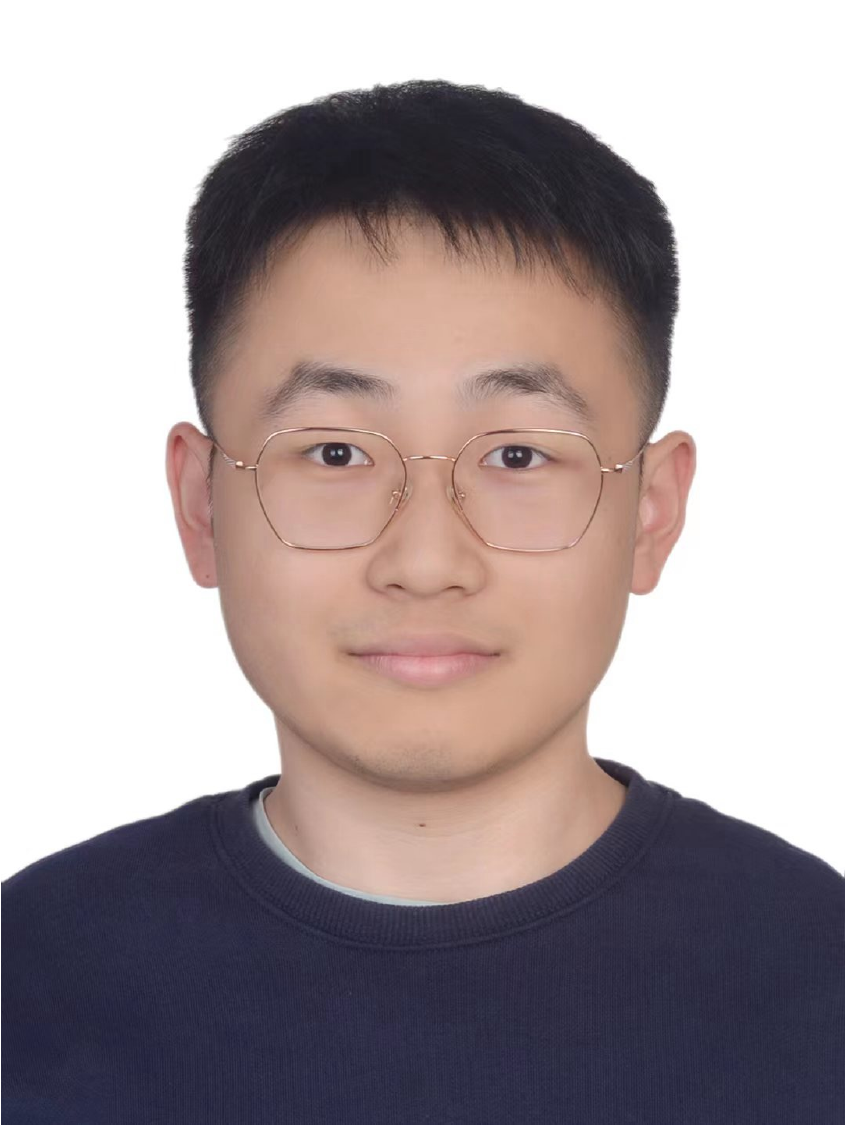}}]{Fangjing Wang}
received a Bachelor's degree in Computer Science and Technology from Dalian University of Technology, Dalian, China, in 2022. He is currently pursuing an M.S. degree in the Department of Computer Science and Engineering at Southern University of Science and Technology (SUSTech), Shenzhen, China. His research interests include computer vision, object tracking, and embodied intelligence.
\end{IEEEbiography}
\vspace{-30pt}
\begin{IEEEbiography}[{\includegraphics[width=1in,height=1.25in,clip,keepaspectratio]{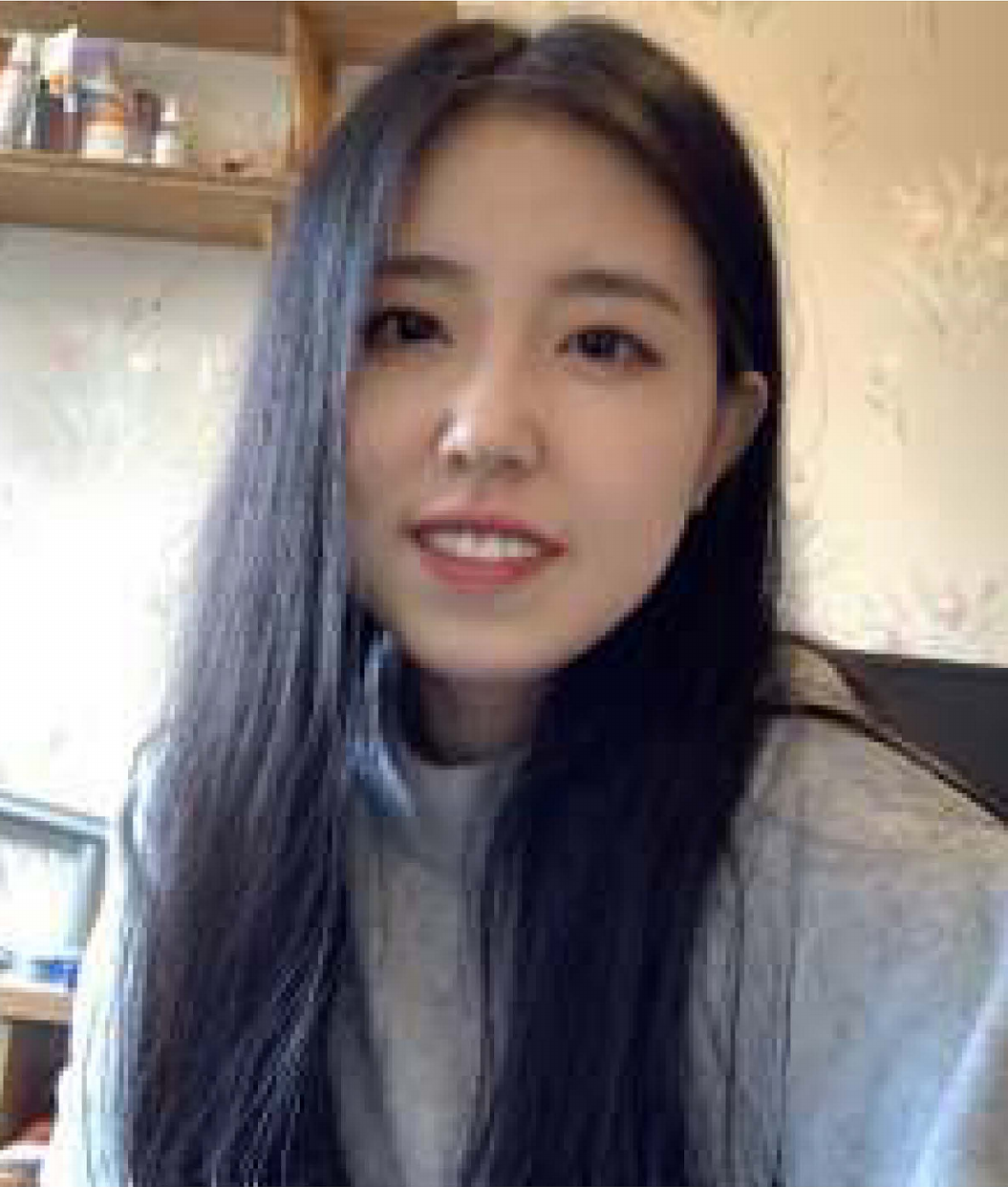}}]{Jinyu Yang}
received a Bachelor's degree in Engineering from Beihang University, Beijing, China, in 2018, an M.S. degree in Electrical and Electronic Engineering from The Hong Kong University of Science and Technology, Hong Kong, in 2019, and a Ph.D. degree from the University of Birmingham and Southern University of Science and Technology (SUSTech) in 2024. She is currently with Harbin Institute of Technology, Shenzhen, China. Her research interests include computer vision and object tracking.
\end{IEEEbiography}
\vspace{-30pt}
\begin{IEEEbiography}[{\includegraphics[width=1in,height=1.25in,clip,keepaspectratio]{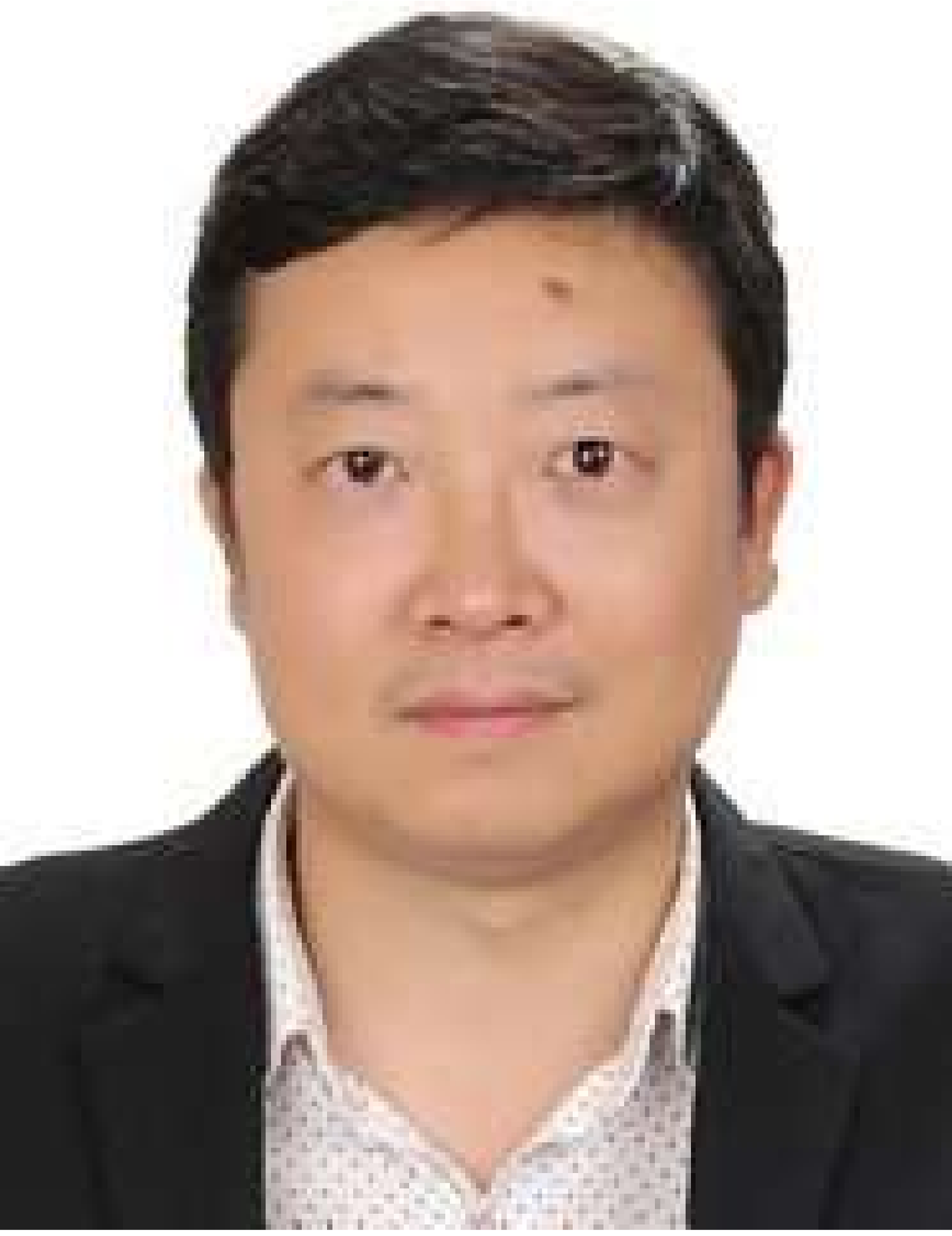}}]{Feng Zheng}
received a Ph.D. degree from the University of Sheffield, UK. Before joining SUSTech, he worked as a Senior Researcher at Tencent YouTu Lab in Shanghai, China. Prior to this, he was a Postdoctoral Researcher at the University of Pittsburgh, USA, and an Assistant Research Professor at the Shenzhen Institute of Advanced Technology, CAS. He is currently an Associate Researcher in the Department of Computer Science and Engineering at Southern University of Science and Technology (SUSTech), Shenzhen, China. His research interests include machine learning, computer vision, and human-computer interaction.
\end{IEEEbiography}

\vfill
\end{document}